\documentclass[twoside]{article}

\usepackage{PRIMEarxiv}

\usepackage[utf8]{inputenc} 
\usepackage[T1]{fontenc}    
\usepackage{hyperref}       
\usepackage{url}            
\usepackage{booktabs}       
\usepackage{amsfonts}       
\usepackage{nicefrac}       
\usepackage{microtype}      
\usepackage{lipsum}
\usepackage{fancyhdr}       
\usepackage{graphicx}       

\usepackage{algorithm}
\usepackage{algpseudocode}

\usepackage{amssymb}
\usepackage{amsmath,amsfonts}
\usepackage{mathrsfs}

\usepackage{multirow}
\usepackage{makecell}
\usepackage{booktabs}
\usepackage{algpseudocode}
\usepackage{siunitx}

\usepackage{caption}
\usepackage{subcaption}

\newcommand{\METHOD}{OPAB}
\newcommand{\METHODSPACE}{OPAB }

\title{Actions Have Consequences: Detecting Outcome Performativity Using Intervention Testing - Extended Version}

\author{
  Brandon Gower-Winter and Georg Krempl \\
  Utrecht University \\
  8 Hiedelberglaan, Utrecht, 3584 CS, NL \\
  \texttt{b.gower-winter@uu.nl, g.m.krempl@uu.nl} \\
}

\begin{document}
\maketitle

\begin{abstract}
In many domains such as Palliative Care, Credit Assignment and Recommender Systems, predictions may causally influence the outcomes they predict. This phenomena is known as Outcome Performativity.
This paper formalises an approach for detecting Outcome Performativity using prediction intervention called Outcome Performativity A/B Detection (\METHOD). \METHODSPACE enables the detection of Outcome Performativity by assessing the dissimilarity in outcome distributions produced by different predictions groups (interventions). If that dissimilarity is significant, Outcome Performativity is detected. 
We derive sample complexity bounds for \METHODSPACE under various Outcome Performative assumption classes which we empirically validate. Results show that detecting Outcome Performativity using \METHODSPACE is achievable in numerous cases.
Results also show the presence of regions of indistinguishability which describe settings where the allotted number of interventions are insufficient for detecting Outcome Performativity. 
The results of which have broader practical implications for the detectability of Outcome Performativity in settings where samples are scarce, cost-prohibitive or potentially unethical to obtain. The paper concludes with a case study on the efficacy of \METHODSPACE on the Open Bandits dataset, and provides directions for future work.
\end{abstract}

\section{Introduction}

\begin{figure}[tb]
\centering
\begin{subfigure}{0.25\textwidth}
    \centering
    \includegraphics[width=\textwidth]{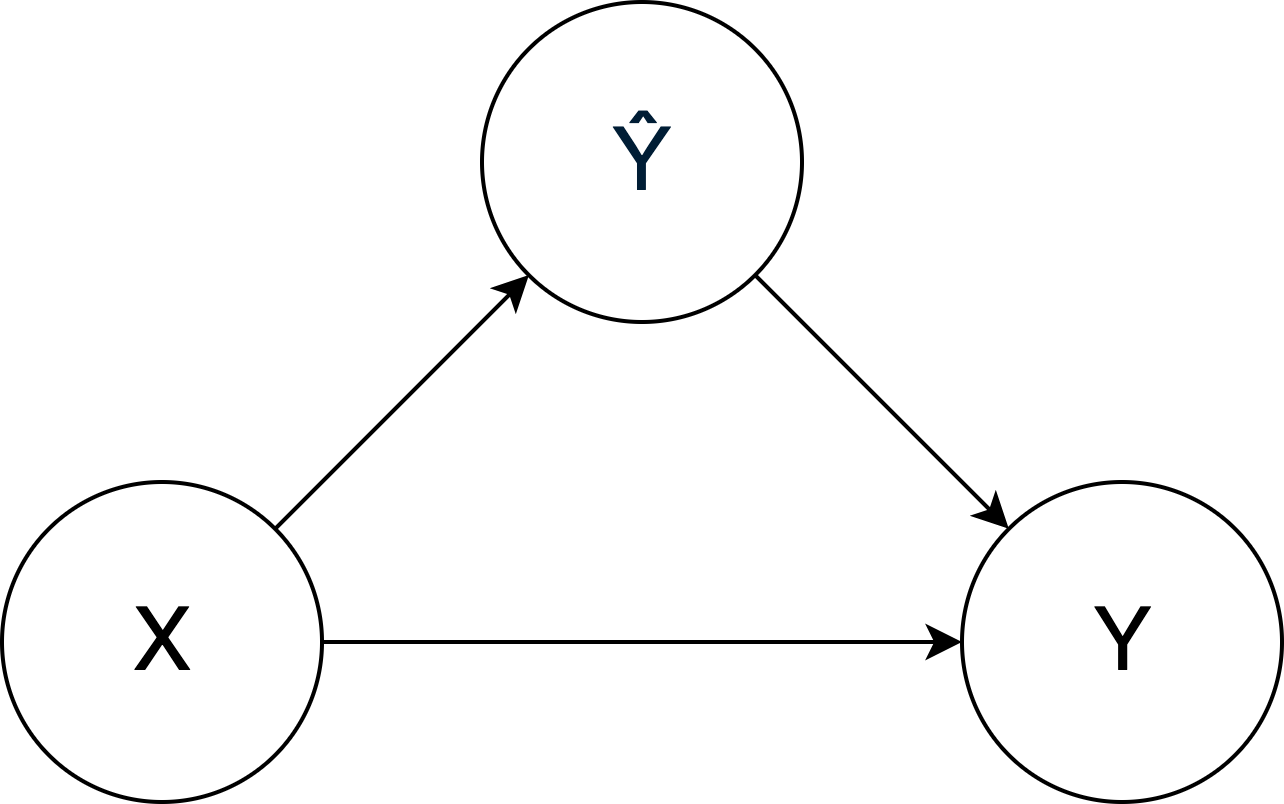}
    \caption{}
    \label{fig:OP_NO_INTERVENTION}
\end{subfigure}
\quad
\begin{subfigure}{0.25\textwidth}
    \centering
    \includegraphics[width=\textwidth]{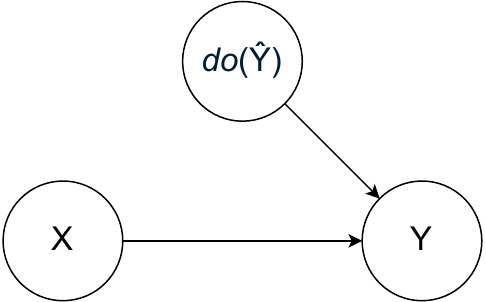}
    \caption{}
    \label{fig:OP_INTERVENTION}
\end{subfigure}
\caption{Causal diagrams of Outcome Performative with (a) no interventions and (b) \METHODSPACE (our method) applied. Here, the outcome (label) $Y$ of a prediction is causally dependent on both the features $X$ and prediction $\hat{Y}$ itself. 
}
\label{fig:OP_CAUSAL}
\end{figure}

It is often assumed that predictions made by Machine Learning (ML) models have no effect on the outcomes they observe, or the data distributions they encounter in the future. However, a growing body of research under the term \textit{Performative Prediction} \cite{PerdomoZrnicMendlerDunner2020} has shown that this is not the case and that many problem domains such as Credit Assignment \cite{ChengHardtMendlerDunner2024} or Patient Care prediction \cite{AdamChangHaibe2020,AmsterdamGelovenKrijthe2025} may exhibit these performative phenomena. In the former, assigning high interest rates to high risk debtors may increase the likelihood that they default on their loans. In the latter, assigning palliative care to a patient (instead of curative care) may ultimately lead to their death.
This work focuses on a specific type of Performativity known as \textit{Outcome Performativity} \cite{KimPerdomo2023} where predictions made by ML models causally affect the outcomes observed (Fig. \ref{fig:OP_NO_INTERVENTION}).

The state-of-art often assumes the Outcome Performativeness of a setting is known \textit{a priori}. However, it has been shown that ML algorithms can induce performative feedback loops inconspicuously and may produce unwanted phenomena such as an increased False Positive Rate (FPR) over time \cite{AdamChangHaibe2020}. This motivates the need to detect such phenomena. If performativity is undetected, practitioners may unknowingly deploy models that exacerbate bias or suffer from performance decay. Despite the existence of online methods for identifying performative effects \cite{ChengHardtMendlerDunner2024,GowerWinterKrempl2025}, there has been little work studying the identifiability of Outcome Performativity in an offline setting (i.e., during the data collection process). If such an offline detection method exists, it would allow practitioners to identify Outcome Performativity before a predictive model is deployed, which may be more ethically and financially  beneficial to an online approach which requires a trained and deployed model.

Taking inspiration from Randomized Controlled Trials (RCTs) \cite{Cartwright2010} and work on A/B (Intervention) testing to detect performative feedback loops \cite{GowerWinterKrempl2025}, we apply intervention testing \cite{Pearl2009} to detect Outcome Performativity (Fig. \ref{fig:OP_INTERVENTION}). By observing the outcome distributions that arise from randomly assigning predictions to instances, we causally assess the dissimilarity of said distributions and, if significant, detect Outcome Performativity. This approach is called \textit{\textbf{O}utcome \textbf{P}erformativity \textbf{A}/\textbf{B} Detection} (\METHOD). In contrast to contemporary methods, \METHODSPACE is an offline method which enables the detection of Outcome Performativity prior to a ML model's training and deployment (i.e., detection occurs during the data labelling process). Given this, we are interested in evaluating the sample efficiency of \METHOD. i.e., the ability to detect Outcome Performativity in as few samples as possible which is critical in settings where label acquisition is impractical. We investigate this by analysing the sample complexity of \METHODSPACE on several assumption classes of Outcome Performativity \cite{AdamChangHaibe2020,KimPerdomo2023}, for which we derive sample complexity bounds. This work can be summarized by the following research questions:

\begin{enumerate}

    \item Under what conditions can intervention testing be used to detect Outcome Performativity in binary classification tasks? For this, we explore three assumption classes of Outcome Performativity in various configurations.

    \item To what degree can the number of instances (sample complexity) needed to detect Outcome Performativity be estimated? This is important for domains where intervention testing is costly or has to be minimized for ethical reasons. 
    
    \item  Using the derived sample complexity bounds, can we identify \textit{regions of indistinguishability}? These are settings where required sample complexity is prohibitive for detecting Outcome Performativity in a practical setting.
    
\end{enumerate}

The rest of the paper follows: Sect. \ref{sect:background} introduces related work, Sect. \ref{sect:methodology} describes the implementation of \METHOD, Sect. \ref{sect:EXPERIMENTAL_DESIGN} derives the sample complexity bounds of \METHODSPACE under different Outcome Performativity assumption classes, and Sect. \ref{sect:discussion} reflects on the strengths and limitations of \METHOD. Sect. \ref{sect:conclusions} concludes the paper. The source code for this work can be found at: \url{https://edu.nl/mhh9d}

\section{Background and Related Work} \label{sect:background}

\textit{Outcome Performativity} was introduced by Mendler-D{\"u}nner et al. \cite{MendlerDunnerDingWang2022} and Kim and Perdomo \cite{KimPerdomo2023}. Outcome Performativity is an extension of Performativity \cite{PerdomoZrnicMendlerDunner2020} whereby the realised outcome for some prediction task is causally dependent on both the features of an instance and the prediction given to that instance by some predictor (e.g., human expert, or ML model). Outcome Performative tasks fall into subset of traditional supervised learning tasks where a prediction must be made before the true outcome is observed. For example, in a credit assignment task, one might want to predict whether an applicant would become insolvent in the near future. In this setting, predicting the likelihood of insolvency may affect whether that state of insolvency is realised. If a loan is granted, the applicant may have enough funds to stay operational in the near future. If the applicant is not granted a loan, that may increase the likelihood that they go bankrupt in that same time period.
Formally, a binary classification Outcome Performative task is described as:

\begin{equation}
    p^* : \mathcal{X} \times \widehat{\mathcal{Y}} \rightarrow [0, 1]
\end{equation}

Here, an outcome $y$ from outcome space $\mathcal{Y} = \{0, 1\}$ is determined by a Bernoulli distribution parametrized by $p^*(x, \hat{y})$ for every instance $x \in \mathcal{X}$ and prediction $\hat{y} \in \mathcal{Y}$ pair. $p^*$, also known as \textit{Nature} \cite{KimPerdomo2023}, describes the true outcome distribution of the system.
In practice, the dynamics of $p^*$ are unknown but some assumption classes exist in literature. Adam et al. \cite{AdamChangHaibe2020,AdamChangHaibe2022} adopt misclassification-based dynamics where an outcome $y$ is only subject to Outcome Performativity if the instance $x$ is misclassified. Their model requires that the notion of a desirable outcome can be described before a prediction is made. In the palliative care problem, it is desirable to offer palliative care to patients who would pass away even if curative interventions were administered because it maximises the quality of the patient's end-of-life care. If a curative intervention would save a patient's life, then it is the desirable outcome because life is preferable to the risk of death. In Mendler-D{\"u}nner et al. \cite{MendlerDunnerDingWang2022}, they use model-based dynamics of $p^*$ which adopts the notion that the Outcome Performative behaviour of instances close together in the feature space should behave similarly when given the same prediction assuming no hidden confounders.

Outcome Performative phenomena are highly relevant in many practical domains. A review from Pagan et al. \cite{PaganBaumannElokda2023} categorized Outcome Performativity under the term \textit{Outcome Feedback Loops} and highlighted the prevalence of Outcome Performativity in both adversarial and non-adversarial domains. Liley et al. \cite{LileyEmersonMateen2021} showed that naive updating of ML models can cause performance degradation in Outcome Performative settings which is particularly relevant in healthcare settings where even an accurate model can produce unwanted phenomena \cite{AmsterdamGelovenKrijthe2025}. Furthermore, Adam et al. \cite{AdamChangHaibe2020,AdamChangHaibe2022} investigated the subtle consequences that arise if practitioners are not aware of the feedback loops induced by their ML models. They show that an over-reliance on the predictions made by a ML model results in an increased False Positive Rate if the model later trains on data it made predictions on.

\section{Methodology} \label{sect:methodology}
Before introducing \METHOD, we motivate why intervention testing is important, and useful, when detecting Outcome Performativity.
In an binary classification Outcome Performative setting, an instance $x$ is given a prediction $\hat{y}$ which results in the observation of some outcome $y$. Formally, we say that outcome $y$ is sampled from a Bernoulli distribution parametrised by $p^*$: $y \sim \text{Bernoulli}(p^*(x, \hat{y}))$ where $p^*$, \textit{Nature}, is defined as a mapping of instances and predictions to some probability of realising an outcome $y=1$:

\begin{equation}
    p^* : \mathcal{X} \times \widehat{\mathcal{Y}} \rightarrow [0, 1]
\end{equation}

While the former definitions neatly capture the underlying mechanisms of Outcome Performativity, they do not describe how one should seek to identify if a setting is Outcome Performative. Given this, we can view Outcome Performativity through the lens of the Bayes Theorem:

\begin{equation}
    P(Y \, | \, X ,\, \widehat{Y}) = \frac{P(\widehat{Y} \, | \, X , \, Y)P(Y \, | X)}{P(\widehat{Y} \, | \, X)}
\end{equation}

where $X$, $\widehat{Y}$, and $Y$ are the observed instances, predictions and outcomes resp. To detect Outcome Performative effects, we look for the following conditions:

\begin{equation}\label{eq:HYPOTHESIS_TEST}
    H(P(Y \, | \, X , \, \widehat{Y} = 0), P(Y \, | \, X , \, \widehat{Y} = 1)) \leq \delta
\end{equation}

where $H$ is a hypothesis test that calculates the dissimilarity of the two distributions given some threshold $\delta$ which controls the type I error rate. Informally, Eq. \ref{eq:HYPOTHESIS_TEST} describes testing to see if the predictions $\widehat{Y}$ in some setting $X$, affect the realised outcomes $Y$. However, this is \emph{not} sufficient to detect Outcome Performativity as we first need to examine from where the predictions $\widehat{Y}$ originate. During outcome data acquisition, or the training of a predictive model, initial predictions about outcomes will come from either a different predictive model or some domain expert $\phi$ (i.e., $\hat{y} = \phi(x)$) Referring back to Eq. \ref{eq:HYPOTHESIS_TEST}, this would mean that our naive test for Outcome Performativity would be:

\begin{equation}\label{eq:HYPOTHESIS_TEST_NAIVE}
    H(P(Y \, | \, X , \, \phi(X) = 0), P(Y \, | \, X , \, \phi(X) = 1)) \leq \delta
\end{equation}

Herein lies the problem. Eq. \ref{eq:HYPOTHESIS_TEST_NAIVE}, at most, allows one to tell if predictor $\phi$'s predictions give rise to different outcome distributions, not if the setting itself is Outcome Performative. This is because the outputs of $\phi$ are themselves dependent on the instances $X$ (Fig. \ref{fig:OP_NO_INTERVENTION}). To remedy this and break the dependency of $\widehat{Y}$ on $X$, we can take inspiration from A/B testing and randomised control trials (RCTs) and perform a do-intervention \cite{Pearl2009} on the predictions themselves. Note that this is equivalent to replacing the expert predictor $\phi$ with a random sampling strategy: $\phi(x) = \mathcal{U}\{0, 1\}$ which decouples the dependency of $\widehat{Y}$ on $X$ (Fig. \ref{fig:OP_INTERVENTION}). This gives us the following:

\begin{equation}\label{eq:HYPOTHESIS_TEST_DO}
    H(P(Y \, | \, X , \, \text{do}(\widehat{Y} = 0)), P(Y \, | \, X , \, \text{do}(\widehat{Y} = 1))) \leq \delta
\end{equation}

Eq. \ref{eq:HYPOTHESIS_TEST_DO} is enough to allow one to detect Outcome Performativity, it however requires the ability to learn some reliable approximation of $P(Y \, | \, X \, , \, \text{do}(\widehat{Y}))$ using methods, such as T-learner \cite{KunzelSekhonBickel2019} or Double ML \cite{ChernozhukovChetverikovDemirer2018}, which are sensitive to intervention imbalances, hyper-parameter and model choices \cite{WangLiZhu2026}. Given that one of the paper's goals is to take the first steps towards estimating the sample complexity bounds for Outcome Performative settings, we make an assumption that the Outcome Performative effects in a given setting are uniform across the feature space. This reduces Eq. \ref{eq:HYPOTHESIS_TEST_DO} to:

\begin{equation}\label{eq:HYPOTHESIS_TEST_FINAL}
    H(P(Y \, | \, \text{do}(\widehat{Y} = 0)), P(Y \, | \, \text{do}(\widehat{Y} = 1))) \leq \delta
\end{equation}

which is achievable by simply performing interventions (i.e., random sampling predictions), and statistically evaluating the outcome distributions $Y$ using an appropriate statistical test such as a Chi-Squared or Fisher's Exact Test. As we show later, the assumption that Outcome Performativity is uniform across the feature space is reasonable for several assumption classes. We do, however, elaborate on the limitations of this assumption in Sect. \ref{sect:discussion}. Sect. \ref{sect:OPAB_motivation} provided an additional perspective on why intervention testing is necessary to detect Outcome Performativity.

\subsection{Outcome Performative A/B Testing}
\par
This section describes the execution logic of Outcome Performative A/B testing (\METHOD). Recall that in order test for Outcome Performativity, we need to perform do-interventions on the predictions $\widehat{Y}$ and observe if there are statistical differences in the outcome distributions of both interventions (Eq. \ref{eq:HYPOTHESIS_TEST_FINAL}). Formally, given a set of $T$ instances $X$ for some binary classification task $Y = \{0, 1\}$ and a significance threshold $\delta$ for controlling the type I error,
\METHODSPACE will randomly assign predictions $\hat{y} \in\widehat{Y} = \{0, 1\} \, \forall \, x \in X$ sampled from a Bernoulli distribution ($\hat{y} \sim \text{Bernoulli}(0.5)$) \cite{KimPerdomo2023} (i.e., do-intervention). This results in two groups of size $N$. For each group of instances $X_{\widehat{Y} = 0}$ and $X_{\widehat{Y} = 1}$ (the A and B groups in A/B testing or the assigned groups in RCTs), \METHODSPACE observes the outcome $y \in Y$ of each instance-prediction pair $(x,\hat{y})$, and then evaluates the dissimilarity of the label distributions $P(Y \, | \, \widehat{Y}= 0)$ and $P(Y \, | \, \widehat{Y}= 1)$ using Eq. \ref{eq:HYPOTHESIS_TEST_FINAL}. If $H(P(Y \, | \, \widehat{Y} = 0), P(Y \, | \, \widehat{Y} = 1)) \leq \delta$, then Outcome Performativity is detected. Alg. \ref{algo:OPAB} provides the pseudocode.

\begin{algorithm}[htbp]
\caption{Pseudocode of \METHOD. Here we assume that $H$ is either a Chi-Squared or Fisher's Exact Test.}
\label{algo:OPAB}
\begin{algorithmic}[1]
    \Require Instances $X$, Function $H$, Threshold $\delta$
    \State $A \gets [0, 0]$ \Comment{Construct frequency tables}
    \State $B \gets [0, 0]$ \Comment{for groups A and B}
    \For{$x \in X$} \Comment{For each instance}
        \State $\hat{y} \gets Bernoulli(0.5)$ \Comment{Assign random prediction}
        \State $y \gets$ observe$(x, \hat{y})$ \Comment{Observe outcome of prediction $\hat{y}$ on instance $x$}
        \If{$\hat{y} = 0$} \Comment{Update frequency tables}
            \State $A[y] \gets A[y] + 1$
        \Else
            \State $B[y] \gets B[y] + 1$
        \EndIf
    \EndFor
    \State \Return $H(A, B) \leq \delta$ \Comment{Return result of hypothesis test}
\end{algorithmic}
\end{algorithm}

\section{Sample Complexity Analysis \label{sect:EXPERIMENTAL_DESIGN}}

Recall that in an Outcome Performative setting, an outcome $y$ is sampled from a Bernoulli distribution parametrised by $p^* : \mathcal{X} \times \widehat{\mathcal{Y}} \rightarrow [0, 1]$ where $p^*$ is \textit{Nature} (i.e., $y \sim \text{Bernoulli}(p^*(x, \hat{y}))$). \textit{Nature} is the true conditional outcome distribution that describes the underlying dynamics of an Outcome Performative setting. In reality, the actual dynamics of $p^*$ are unknown, but several assumption classes describing $p^*$ exist. Each subsequent subsection derives the sample complexity bounds for three of those assumption classes and empirically validates them using \METHOD. Deriving sample complexity bounds for Outcome Performative settings is of practical importance because it allows practitioners to calculate the minimum sample size needed to reliably detect Outcome Performativity, which is vital in high-stakes domains like healthcare where the cost of intervention testing is either ethically ambiguous or prohibitively expensive. 

In order to empirically evaluate two of the assumption classes: Model-based (Sect. \ref{sect:MBOP}) \cite{MendlerDunnerDingWang2022}, and Misclassification-based (Sect. \ref{sect:ADAM_PERF}) \cite{AdamChangHaibe2020}, real-data which is imputed with Outcome Performativity is required. The datasets used in this work are: \verb|breast| \verb|cancer| \cite{DATASET_BREAST_CANCER}, \verb|diabetes| \cite{DATASET_DIABETES}, \verb|adult| \verb|census| \cite{DATASET_ADULT_CENSUS}, \verb|kickstarter| \cite{DATASET_KICKSTARTER}, \verb|titanic| \cite{DATASET_TITANIC}, and \verb|loan| \cite{DATASET_LOAN}. They are all binary classification tasks. 
Unless stated otherwise, results reported in this section for every parameter combination are averaged over $100$ replicates and a pseudorandom number generator is used to ensure reproducibility. We use the Chi-Squared Test to detect Outcome Performativity and $\delta=0.05$.
A description of each dataset, the source code for all experiments, and additional experimental results such as using different statistical tests to determine significance or varying the parameter $\delta$, are can be found in the Appendix.

\subsection{Simple Outcome Performativity}\label{sect:SOP}

The first assumption class we investigate is a simplified model of Outcome Performativity. In this setting, we drop $p^*$'s dependence on the instance space $\mathcal{X}$ describing it as follows:

\begin{equation}
    p^* = \begin{cases}
        \alpha_0 & \verb|if | \hat{y} = 0 \\
        \alpha_1  &\verb|if | \hat{y} = 1 \\
    \end{cases}
\end{equation}

where $\alpha_c = P(y=1 \, | \, \hat{y} = c)$ is the probability that the outcome is $y=1$ when the prediction is $\hat{y} = c \in \{0, 1\}$ (binary classification task). In this setting, a realized outcome $y$ is only dependent on the prediction $\hat{y}$. We acknowledge that this assumption class is unrealistically simple, but we include it because it allows us identify key behavioural dynamics of Outcome Performativity as well as derive sample complexity bounds which are used as building blocks for deriving sample complexity bounds for more sophisticated Outcome Performative settings.

In order to derive sample complexity bounds, we need to make assumptions about the hypothesis test $H$ and the significance threshold $\delta$,  Assuming $H$ is a Chi-Squared Test, the estimated minimum number of interventions per prediction group to detect Outcome Performativity is given by the following Eq. (see Sect. \ref{sect:proofs} for proof):

\begin{equation} \label{eq:N_APPROX_SIMPLE}
    N = \chi_\delta^2 \, \frac{2(\alpha_0 + \alpha_1) - (\alpha_0 + \alpha_1)^2}{2(\alpha_0 - \alpha_1)^2}
\end{equation}

where $\chi_\delta^2$ is the critical $\chi^2$ value for some significance threshold $\delta$ (e.g., this work uses $\delta =  0.05$, therefore the $\chi_{\delta=0.05}^2$ with one degree of freedom is $3.841$).
Eq. \ref{eq:N_APPROX_SIMPLE} reveals several important findings: the first is that fewer instances are needed to detect Outcome Performativity as $|\alpha_0 - \alpha_1| \to 1.0$. This makes sense as the resulting outcome distributions will be increasingly different as the performative response of $\alpha_0$ and $\alpha_1$ differ greatly. Secondly, the number of instances required per intervention $N \to \infty $ when $\alpha_0 \approx \alpha_1$. This means that for arbitrarily small differences in Outcome Performative responses per intervention, detecting Outcome Performativity becomes increasingly prohibitive. 

From a practical perspective, the derivation of sample complexity bounds for an Outcome Performative setting introduces another concept: \textit{regions of indistinguishability}. Formally, given an allotted budget per intervention $\bar{N}$, confidence threshold $\delta$, and assumption class $\mathcal{A}$, the \textit{regions of indistinguishability} are defined as the set of parametrisations of $a \in \mathcal{A}$ where the allotted interventions per group are less than the minimum required number of instances defined by the complexity bound $N_{\delta, a}$:

\begin{equation} \label{eq:ROI}
    \{a \, | \, a \in \mathcal{A} \text{ and } \bar{N} < N_{\delta, a} \}
\end{equation}

Note that $N_{\delta, a}$ is proxy notation for a sample complexity bound calculation. For Simple Outcome Performativity, $N_{\delta, a}$ is Eq. \ref{eq:N_APPROX_SIMPLE} and $a$ is some 2-tuple that parametrises the setting (i.e., $a = (\alpha_0, \alpha_1)$). However, the concept of the \textit{regions of indistinguishability} in an Outcome Performative setting are both assumption class, and sample complexity bound agnostic.  

\begin{figure}[tb]
     \centering
     \includegraphics[width=\textwidth]{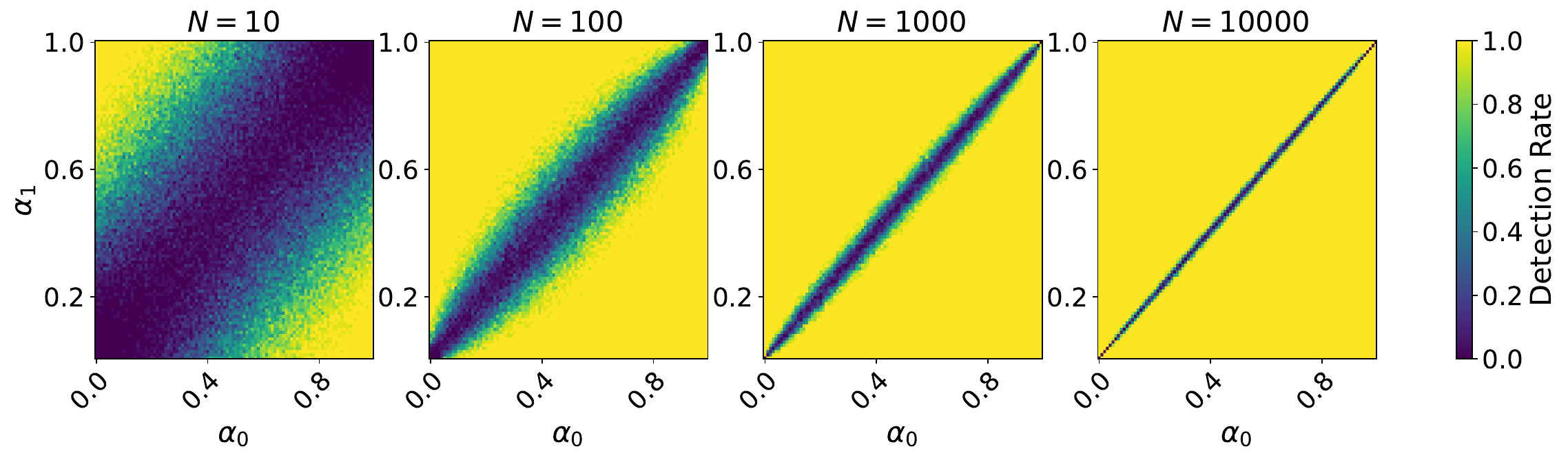}
    \caption{shows the detection rate of \METHODSPACE in the Simple Outcome Performativity setting as the allotted interventions per group $N$ increases. Dark blue values indicate \textit{regions of indistinguishability} where, for the given $N$, Outcome performativity cannot be reliably detected.}
    \label{fig:SIMPLE_MODEL_N}
\end{figure}

\subsubsection{Results}

To empirically validate these findings, we run simulations of a Simple Outcome Performative setting over various $\alpha_0$ and $\alpha_1 \in [0,1]$ using \METHODSPACE ($\delta = 0.05$) to detect the presence of Outcome Performativity. Fig. \ref{fig:SIMPLE_MODEL_N} reports the detection rate of \METHODSPACE over values of $\bar{N} = \{10, 100, 1000, 10000\}$. The detection rate reported in each cell is the number of times \METHODSPACE was able to detect Outcome Performativity divided by the total number of repeated runs (25 in this case). These results confirm the presence of \textit{regions of indistinguishability} (Fig. \ref{fig:SIMPLE_MODEL_N}, dark blue regions): As $\bar{N}$ increases, the number of parameter combinations ($\alpha_0$, $\alpha_1$) decreases proportionally. In the simple Outcome Performativity setting, when $|\alpha_0 - \alpha_1|$ is large, few instances ($\bar{N}=10$) are required to reliably detect Outcome Performativity. However, when $\alpha_0 \approx \alpha_1$, even a large amount of instances ($\bar{N}=10\,000$) cannot detect Outcome Performativity.
Interestingly, despite assuming a Chi-Squared test is used to detect Outcome Performativity, the estimated $N$ produced by Eq. \ref{eq:N_APPROX_SIMPLE} is applicable when using other statistical tests to detect Outcome Performativity (Results using Fishers Exact Test and Mann-Whitney U Test are shown in Sect. \ref{sect:additional_hypothesis_tests}).

\subsection{Model-based Outcome Performativity} \label{sect:MBOP}

\begin{figure}[tb]
\centering
\includegraphics[width=0.4\textwidth]{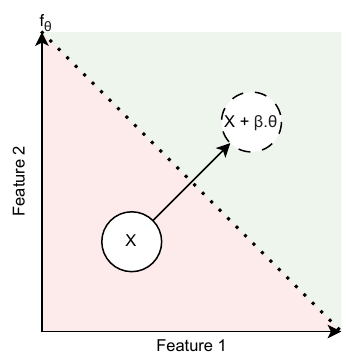}
\caption{A visualization of the Model-based Outcome Performativity Model on a two dimensional feature space. $f_{\theta}$ is a linear model trained on a labelled dataset and its decision boundary is represented by the dotted line. The red region indicates where $f_{\theta}(x) = 0$ and the green region indicates where $f_{\theta}(x) = 1$. To simulate Outcome Performativity an instance $x$ is moved in the feature space using the linear model's weights $\theta$ and Outcome Performative strength $\beta$. The result of $f_{\theta}(x + \beta.\theta)$ is then the observed outcome $Y$. In this example, the instance would initially have an observed outcome of $Y=0$, but because of the simulated Outcome Performativity, the observed outcome is $Y=1$ instead.}
\label{fig:FEATURE_DEPENDENT_VISUAL}
\end{figure}

Recall that in Outcome Performative settings, the outcome of an instance-prediction pair is determined by $p^* : \mathcal{X} \times \widehat{\mathcal{Y}} \rightarrow [0, 1]$. The previous assumption class abstracted away $\mathcal{X}$. This section reintroduces $\mathcal{X}$ with Model-based Outcome Performativity. In Model-based Outcome Performativity \cite{MendlerDunnerDingWang2022}, an instance's position in the feature space is taken into consideration before applying any Outcome Performative Effects. More specifically, Model-based Outcome Performativity assumes that $p^*$ can be represented by a linear model $f_{\theta} \to \{0, 1\}$ and its weights $\theta$. Using $f_\theta$ and $\theta$, Outcome Performativity is modelled as:

\begin{equation} \label{eq:model_based_nature}
    p^* = f_{\theta}(x + \beta_{\hat{y}}.\theta)
\end{equation}

where $\beta_{\hat{y}}$ is the strength of the Outcome Performativity for prediction $\hat{y} = c$. If $\beta_c < 0.0$, it biases instances towards an outcome of $y = 0$. If $\beta_c > 0.0$, it biases instances towards an outcome of $y = 1$. If $\beta_c = 0.0$, predictions are non-Outcome Performative. A visualization of Model-based Outcome Performativity is shown in Fig. \ref{fig:FEATURE_DEPENDENT_VISUAL}.
Sample complexity analysis shows that Model-based Outcome Performativity behaves similarly to Simple Outcome Performativity (Eq. \ref{eq:N_APPROX_SIMPLE},) despite the introduction of the feature space $\mathcal{X}$ (see Sect. \ref{sect:proofs} for proof):

\begin{equation} \label{eq:N_APPROX_FEATURE}
    N = \chi^2_\delta \, \frac{ 2(p_{\beta0} + p_{\beta 1}) - (p_{\beta 0} + p_{\beta 1})^2}{2(p_{\beta 0} - p_{\beta 1})^2}
\end{equation}

where $p_{\beta c} = P(Y=1 \, | \, \widehat{Y} = c \, , \, \theta)$ is the proportion of instances that would have an outcome of $y=1$ given a prediction of class $\hat{y} = c$ when using model $f_\theta$. When $|\beta_0 - \beta_1| \to\infty$, the total number of instances $N$ needed per intervention group $\to 0.0$. Conversely, when $|\beta_0 - \beta_1| \to 0.0$, $N \to \infty$. The caveat is that the rate at which this happens depends on the parametrisation of $f_\theta$, which we show in the next section.



\subsubsection{Results}

In order to simulate Model-based Outcome Performativity, a predictive model $f_\theta$ and its parameters $\theta$ are needed. We achieve this by using non-Outcome Performative binary classification dataset (e.g., \verb|Diabetes|), and train $f_\theta$ on that dataset. This enables one to impute the non-Outcome Performative dataset with Model-based Outcome Performativity. This approach preserves class imbalances, and feature densities from the original non-Outcome Performative dataset. In order to simulate Outcome Performativity, one samples an instance $x$ from the original dataset, assigns a prediction $\hat{y}$ and applies Eq. \ref{eq:model_based_nature} to observe the final outcome $y$. To evaluate the sample complexity bounds derived in Eq. \ref{eq:N_APPROX_FEATURE}, we evaluate the detection rate of \METHODSPACE on the datasets described in Sect. \ref{sect:EXPERIMENTAL_DESIGN} imputed with Model-based Outcome Performativity. When evaluating \METHODSPACE we randomly sample $10\%$ of the instances from each dataset without replacement. We do this because it is unreasonable in many cases to assume that some large percentage of the total available instances would undergo intervention testing (i.e., intervention testing could be cost prohibitive or unethical at large scales).

\begin{figure}[tb]
     \centering
     \includegraphics[width=\textwidth]{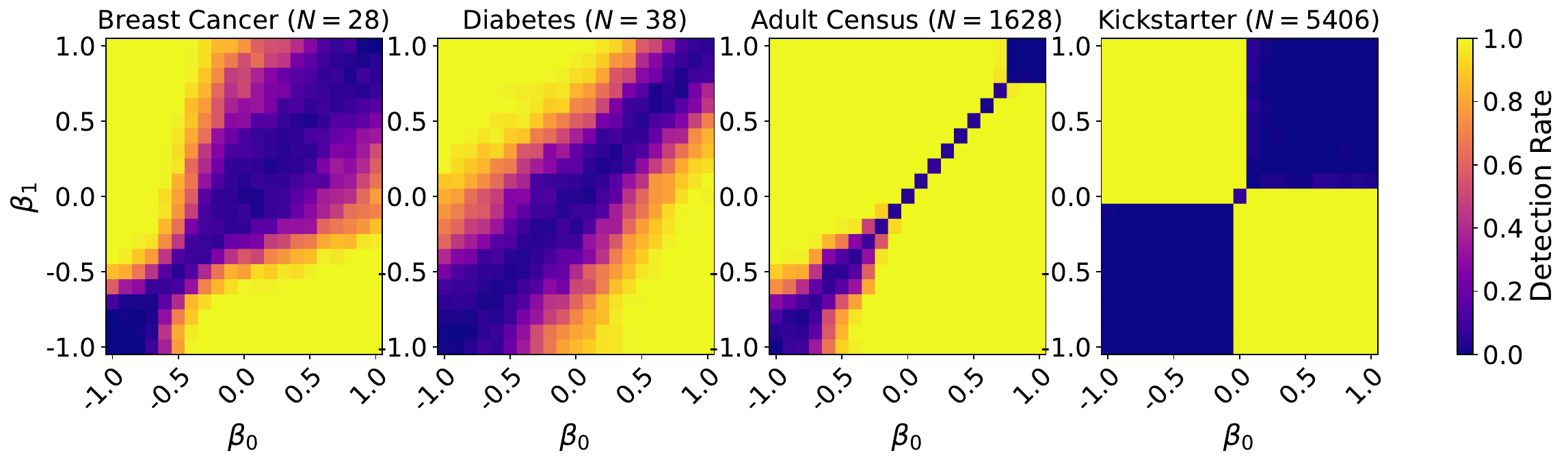}
    \caption{shows the detection rate of \METHODSPACE in Model-based Outcome Performativity settings for different parameter combinations ($\beta_0$, $\beta_1$) over several datasets. Dark blue parameter combinations reveal \textit{regions of indistinguishability} for given datasets given the allotted interventions per group $N$.}
    \label{fig:FEATURE_DEPENDENT_MODEL}
\end{figure}

Fig. \ref{fig:FEATURE_DEPENDENT_MODEL} illustrates the results of this experiment where we find that as $N$ increases, the detection rate of \METHODSPACE increases. This is most clearly shown across the \verb|breast| \verb|cancer| and \verb|adult| \verb|census| datasets where for the latter, \METHODSPACE reliably detects Outcome Performativity across a range of $\beta_0$ and $\beta_1$ parameter values. For the former, the range of $\beta_0$ and $\beta_1$ parameters that \METHODSPACE detects Outcome Performativity for decreases (see Sect. \ref{sect:feat_vary_N}).
Results also confirm non-uniform Outcome Performativity responses across the datasets. The \verb|breast| \verb|cancer| dataset illustrates this where a class imbalance in the training dataset (bias towards $Y=1$) causes non-uniformity in the number of interventions needed per class $N$. As $\beta_c \to 1.0$ (the biased class), $N$ increases as the outcome distributions rapidly converge. The original \verb|breast| \verb|cancer| dataset is imbalanced favouring benign samples ($Y = 1$) which means that as $\beta_0$ and $\beta_1 \to 1.0$, $f_\theta$ more rapidly produces identical distributions across both prediction groups. This effect is seen less intensely in the \verb|diabetes| setting where the original dataset is slightly imbalanced favouring $Y = 0$.
These results further support the presence of \textit{regions of indistinguishability} (Fig. \ref{fig:FEATURE_DEPENDENT_MODEL}, dark blue regions). Unlike the Simple Outcome Performative setting (Sect. \ref{sect:SOP}), Model-based Outcome Performative settings reveal how imbalances in the Outcome Distributions and feature space density can significantly increase the number of instances required to detect Outcome Performativity. This is shown in Fig. \ref{fig:FEATURE_DEPENDENT_MODEL} where certain combination of ($\beta_0$, $\beta_1$) are trivially easy to detect, but others are practically infeasible with the allotted interventions ($\bar{N}$).

\subsection{Misclassification-based Outcome Performativity}\label{sect:ADAM_PERF}

The final assumption class we evaluate is Misclassification-based Outcome Performativity \cite{AdamChangHaibe2020,AdamChangHaibe2022} where $p^*$, \textit{Nature}, is described as:

\begin{equation} \label{eq:MBOP}
    p^* = \begin{cases}
        1.0 & \verb|if | \hat{y} = 1 \verb| and | \bar{y} = 1 \\
        0.0 & \verb|if | \hat{y} = 0 \verb| and | \bar{y} = 0 \\
        \gamma_1 & \verb|if | \hat{y} \neq 1 \verb| and | \bar{y} = 1 \\
        1 - \gamma_0 & \verb|if | \hat{y} \neq 0 \verb| and | \bar{y} = 0 \\
    \end{cases}
\end{equation}

Unlike the previous two assumption classes, Misclassification Outcome Performativity only occurs when some prediction $\hat{y}$ is not equal to some desirable outcome $\bar{y} =c$. What constitutes a desirable outcome is domain specific. In the palliative care problem, it is desirable to give curative care to patients who are likely to survive, whereas palliative is desirable when patients will pass away regardless of any type of medical intervention. In the credit assignment problem, it is desirable to grant credit to entities who will be able to pay the debt back. 

In Misclassification Outcome Performativity, when a mismatch between a prediction $\hat{y}$ and desirable outcome $\bar{y}$ occur, the probability the desired outcome is realized despite misclassification (i.e., $y=c$) is defined by the probability $\gamma_c \in [0, 1]$. When $\gamma_c = 0$, desired outcomes are never realized upon misclassification and when $\gamma_c = 1$, the desired outcome is always realized despite misclassification (i.e. no Outcome Performativity).

\par
To derive sample complexity bounds for Misclassification  Outcome Performativity, we again assume $H$ is the Chi-Squared Test. The estimated number of interventions needed per class $N$ is:

\begin{equation}\label{eq:N_APPROX_ADAM}
    N = -\chi^2_\delta \, \frac{(\gamma_0p_0 +\gamma_1p_0 - \gamma_1 - 1)(\gamma_0p_0 +\gamma_1p_0 - \gamma_1 + 1)}{2 \times (\gamma_0p_0 -\gamma_1p_0 + \gamma_1 - 1)^2}
\end{equation}

Where $p_0$ is the probability of encountering a desired outcome $\bar{y} = 0$ when randomly selecting an instance and desired outcome pair $(x, \bar{y})$. See Sect. \ref{sect:proofs} for proof deriving Eq. \ref{eq:N_APPROX_ADAM} and visual aid for different $p_0$ estimates. Eq. \ref{eq:N_APPROX_ADAM} shows that as $\gamma_0$ and $\gamma_1 \to 1.0$, $N \to \infty$ meaning that weaker Outcome Performativity is harder to detect.
This makes sense because as $\gamma_c \to 0.0$, more instances will not realize their desired outcomes when misclassified which creates increasingly different outcome distributions. Conversely, as $\gamma_c \to 1.0$ fewer instances will not realize their desired outcomes resulting in increasingly similar outcome distributions.

\subsubsection{Results}

We empirically validate Eq. \ref{eq:N_APPROX_ADAM}, by applying \METHODSPACE to the datasets described before imputed with Misclassification Outcome Performativity. To achieve this, the set of desirable outcomes $\overline{Y}$ is determined by the dataset's original labels prior to being imputed with Outcome Performativity. If a prediction mismatches the desired outcome, we simulate Misclassification Outcome Performativity using Eq. \ref{eq:MBOP}. Fig. \ref{fig:ADAM_MODEL} shows the results of four datasets (others are included in the appendix). We allocate $10\%$ of the total dataset for detecting Outcome Performativity with each instance being randomly sampled without replacement.

\begin{figure}[tb]
     \centering
     \includegraphics[width=\textwidth]{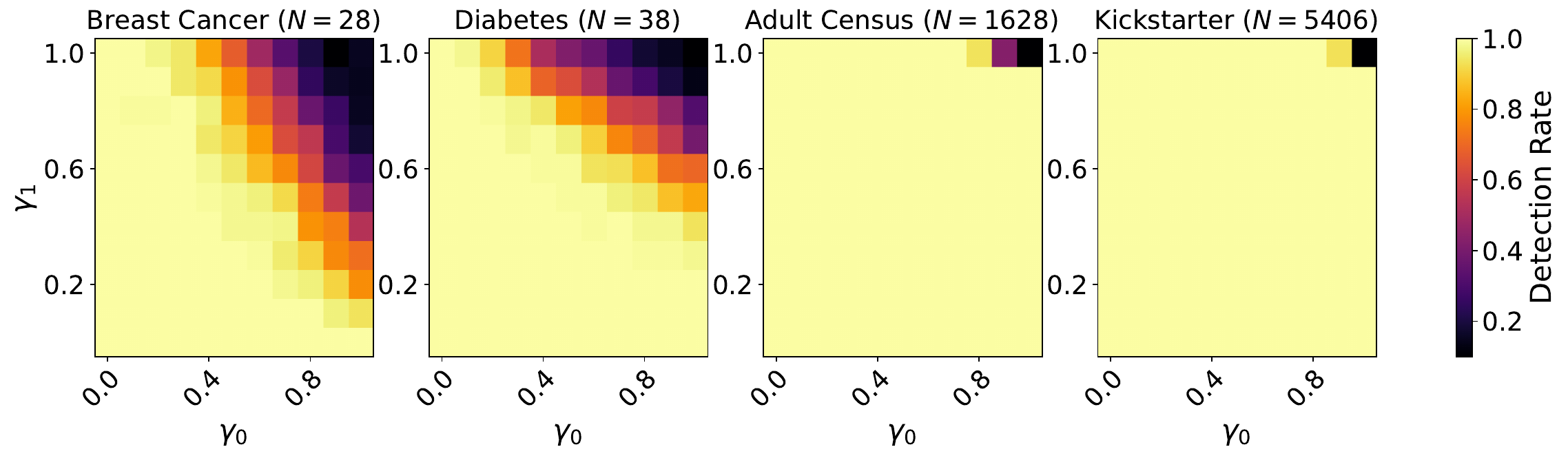}
    \caption{shows the detection rate of \METHODSPACE in Misclassification-based Outcome Performativity settings. Black cells indicate parameter combinations ($\gamma_0$, $\gamma_1$) in the \textit{regions of indistinguishability} for the allotted interventions per class $N$.}
    \label{fig:ADAM_MODEL}
\end{figure}

Empirical results confirm the theoretical findings of Eq. \ref{eq:N_APPROX_ADAM}. When $N$ is large, \METHODSPACE can detect Outcome Performativity over most $\gamma_0$ and $\gamma_1$ values investigated. The exception being when $\gamma_0 =\gamma_1 = 1.0$ which is expected as this parameter combination is non-Outcome Performative. Interestingly, \textit{regions of indistinguishability} (Fig. \ref{fig:ADAM_MODEL}, black regions) seem less common in the Misclassification-based Outcome Performativity setting. This is expected because there is only one parameter combination ($\gamma_0 = 1.0$, $\gamma_1 = 1.0$) which is truly Outcome Performative as opposed to the infinite parameter combinations in the Simple ($\alpha_0 = \alpha_1$) and Model-based ($\beta_0 = \beta_1$) Outcome Performative settings. Given this, our analyses suggest that detecting Misclassification-based Outcome Performativity (using \METHODSPACE or otherwise) requires fewer interventions than the other assumption classes. 

\section{Case Study: Open Bandits Dataset}\label{sect:case_study}

To the best of our knowledge, no dataset exists which explicitly studies Outcome Performativity. However, Outcome Performativity can be inferred if there are instances for which (1) predictions are recorded and (2) these predictions are randomly assigned. The Open Bandits Dataset (OBD) \cite{SaitoShunsukeMegumi2020} meets these criteria. OBD was constructed using multi-armed bandits on the fashion e-commerce platform ZOZOTOWN for the off-policy evaluation of recommender systems. Each instance represents a user impression containing feature values, item recommendations and their placement (clothing items to choose from and where they appear on the store webpage), and click indicators (was the recommended clothing item clicked on) as an outcome. For a subset of the data, the placement of each recommended item on the store page (left, centre, or right) was determined using random sampling. We use this subset and repurpose the placement of item recommendation as predictions $\widehat{Y}$ and click indications as outcomes $Y$.

The dataset is also divided into sub-datasets for both \verb|men| and \verb|women|. Applying OP-AB to all the instances in both of these sub-datasets we find Outcome Performativity in \verb|men| ($p=0.014$), but not in \verb|women| ($p=0.87$). We cannot speak to gender-based purchasing habits, but these results would indicate that the order in which recommended items appear on the ZOZOTOWN platform are not necessarily relevant for women (not outcome performative), but are relevant for men (outcome performative). Nevertheless, we use the \verb|men| dataset to study the efficacy of OP-AB  and the \verb|women| dataset to study the prevalence of false positives produced by OP-AB across various $N$. Sect. \ref{sect:experimental_design} give a complete description of the dataset and all pre-processing steps.

\begin{table}[tb]
    \centering
    \caption{Detection Rate  of OP-AB across varying sample sizes ($N$) on the OBD datasets. These results show that OP-AB has a low false detection rate as shown on the non-Outcome performative \textit{women} dataset and a increased efficacy as $N$ increases as shown the Outcome Performative \textit{men} dataset}
    \label{tab:OBD_unbalanced}
    \begin{tabular}{lccccc}
        \toprule
        dataset / $N$ & $10$ & $10^2$ & $10^3$ & $10^4$ & $10^5$ \\
        \midrule
        women & 0.0 & 0.001 & 0.042 & 0.042 & 0.016 \\
        men   & 0.0 & 0.0   & 0.05  & 0.094 & 0.523 \\
        \bottomrule
    \end{tabular}
\end{table}

We first evaluated OP-AB ($\delta = 0.05)$ on the OBD datasets by taking random samples (without replacement) of size $N = \{10, 10^2, 10^3, 10^4, 10^5\}$ over $1000$ replicates. The detection rate reported in Table \ref{tab:OBD_unbalanced} shows that OP-AB has a low false positive of about $[0.0,0.05]$ in the \verb|women| dataset. Conversely, OP-AB seemingly struggles to detect Outcome Performativity in the \verb|men| dataset only being $50\%$ accurate at the largest $N = 10^5$. This occurs for two reasons: (1) the Outcome Performative effect in this dataset is quite small and thus harder to detect which requires exponentially more instances with random predictions. This claim is supported by the low Phi coefficient ($\Phi = 0.004$) which indicates a negligible effect size and is a limitation of this dataset (2) the class imbalance of the dataset makes it harder for Outcome Performative effects to be observed. The OBD datasets are heavily imbalanced with only about $0.05\%$ of all item recommendation resulting in a click in our setup.

\begin{table}[tb]
    \centering
    \caption{Detection Rate of OP-AB across varying sample sizes ($N$) on balanced variants of the OBD datasets. The results show that OP-AB is more sample efficient when the outcome distribution is balanced. However, the detectability of Outcome Performativity in ultimately dependent on performative strength (effect size) which is negligible in these datasets.}
    \begin{tabular}{lccccc}
        \toprule
        dataset / $N$ & $10$ & $50$ & $100$ & $500$ & $1000$ \\
        \midrule
        women & 0.046 & 0.059 & 0.06 & 0.051 & 0.032 \\
        men   & 0.044 & 0.068 & 0.074  & 0.159 & 0.234 \\
        \bottomrule
    \end{tabular}
    \label{tab:OBD_balanced}
\end{table}

To investigate the potential class imbalance further, we created balanced datasets of both the \verb|women| and \verb|men| datasets by randomly downsampling the majority class (without replacement) over $1000$ replicates and repeating the experiment over $N = \{10, 50, 100, 500, 1000\}$ (sample size $N$ must decrease as the total dataset decreases in size). The detection rates are reported in Table \ref{tab:OBD_balanced} and confirm that OP-AB is more effective in settings where the outcome distribution $P(Y)$ is balanced. For example, OP-AB was four times more effective at detecting Outcome Performativity when $N=10^3$ on the balanced \verb|men| dataset.

\section{Discussion} \label{sect:discussion}

This work formalised the act of performing intervention testing on predictions in order to detect Outcome Performativity. This method, \METHOD, was evaluated on several theoretical models (assumption classes) of Outcome Performativity for which we also derived sample complexity bounds (i.e., the minimum number of interventions per class needed to detect Outcome Performativity). To the best of our knowledge, there exist no other works which explicitly derive sample complexity bounds for detecting Outcome Performativity. Given this, we believe our work provides a practical contribution to the state-of-the-art. The implications of these findings are threefold:

First, by formally describing the process of performing intervention testing on predictions for binary classification tasks, \METHODSPACE enables practitioners to detect Outcome Performativity in real-world domains. This contrasts with related works which have only looked at the identifiability of Outcome Performativity in purely theoretical or semi-synthetic domains. To support the claim that \METHODSPACE can be used to detect Outcome Performativity in real-world settings, we conducted a case study on the Open Bandits Dataset \cite{SaitoShunsukeMegumi2020}.

Second, \METHODSPACE is an offline algorithm, This enables practitioners to detect Outcome Performativity before a model is deployed which can be more cost effective and prevent unwanted performative phenomena from arising once a model is deployed (Appendix includes a demonstration of why training a classifier in an Outcome Performative setting is different than in the classical setting). The complexity analysis of Eq.s \ref{eq:N_APPROX_SIMPLE}, \ref{eq:N_APPROX_FEATURE}, and \ref{eq:N_APPROX_ADAM} provide guidance on the number of instances needed per intervention group.

Lastly, the derivation of sample complexity bounds provides practical insight into the general applicability of intervention testing in various domains. In particular, the introduction of \textit{regions of indistinguishability} which describe Outcome Performative settings for which the allotted number of interventions cannot reliably detect Outcome Performativity. Unsurprisingly, more interventions are needed to detect subtler Outcome Performative effects. This suggests that there may be settings for which detecting Outcome Performativity is practically infeasible. The consequences of such a finding are domain-specific. If the Outcome Performative effects of some setting are subtle (requiring a large number of intervention), but interventions are costly or unethical, it may be inappropriate to deploy ML models in that domain unless other, non-interventional methods, can be used to detect Outcome Performativity. The caveat being that non-interventional methods tend to be less sample efficient than intervention tests \cite{WangLiZhu2026} which poses additional challenges around data availability.

This work has several limitations. First we did not consider the verification latency \cite{KremplZliobaiteBrzezinskiEtal2014} that might occur before an outcome is realized. In such settings, one would either have to wait until all the instances given to \METHODSPACE are realized (which could take a considerable amount of time) 
or the hypothesis testing would need to be done with the subset of instances whose outcomes have been realized. Our work also assumes that the labels recorded when observing an outcome are reliable (i.e. no noisy labels \cite{RenWangZhang2023}). If both of these assumptions do not hold, it may increase the chance that \METHODSPACE produces a type I or type II error. 
Lastly, this work assumes that detecting Outcome Performativity is reliably achievable by observing the difference in outcome distributions of each group of interventions (i.e., \METHODSPACE tests if $P(Y| \text{ do}(\widehat{Y} = 0)) \neq P(Y| \text{ do}(\widehat{Y} = 1))$). If this assumption does not hold for a given setting where Outcome Performativity is only detectable through $P(Y| \, X, \text{ do}(\widehat{Y} = 0)) \neq P(Y| \, X,\text{ do}(\widehat{Y} = 1))$, \METHODSPACE is able to identify if a setting is Outcome Performative, but it cannot identify is a setting is \emph{not} Outcome Performative. \METHODSPACE is sensitive to non-uniform Outcome Performative behaviour. However, this work has shown that \METHODSPACE and more specifically analysis of the outcome distributions under prediction interventions can reliably detect Outcome Performativity in a myriad of settings. Future work will address this limitation by relaxing the uniformity assumption.

\section{Conclusions and Future Work} \label{sect:conclusions}
In this work we formalise the process of detecting Outcome Performativity using prediction interventions. This method: Outcome Performativity A/B Detection (\METHOD) is empirically evaluated across several models of Outcome Performativity for which we also derive sample complexity bounds. This work also introduces the concept of \textit{regions of indistinguishability} which describe Outcome Performative settings where given an allotted number of interventions, Outcome Performativity cannot be reliably detected. Broadly speaking, maximising the number of interventions minimises \textit{regions of indistinguishability}. These results have implications for settings where interventions are cost prohibitive or potential unethical. Future work will look to address the limitations of this paper. Namely, settings with verification latency (when outcomes take some amount of time to be observed) and imperfect / noisy labelling will be considered.

\appendix

\section{OP-AB Motivation}\label{sect:OPAB_motivation}

\begin{figure}[tbp]
    \centering
    \includegraphics[width=0.8\linewidth]{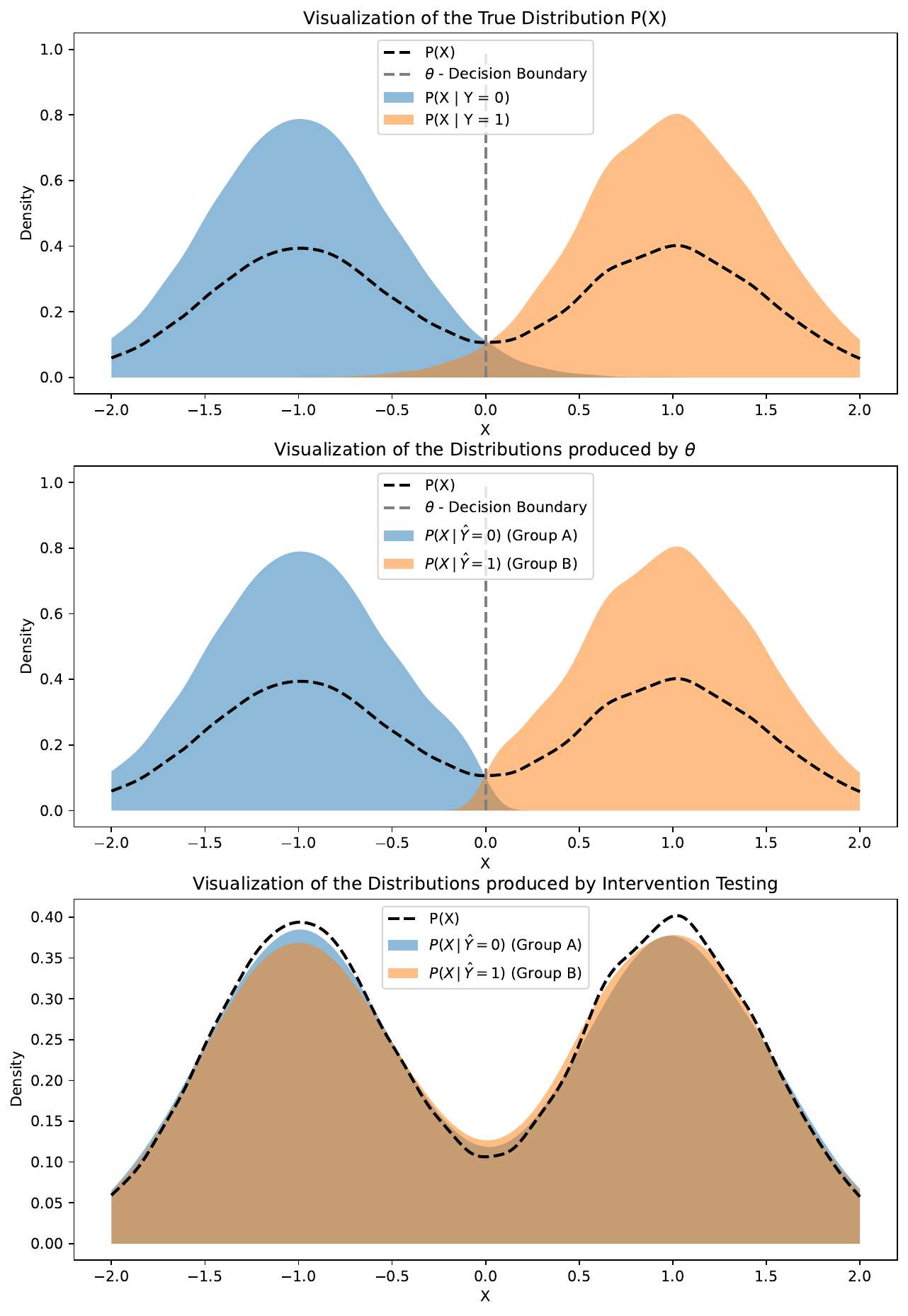}
    \caption{A visualization of the motivation behind OP-AB. The first row shows the true distribution $P(X)$ of an example binary classification dataset $X$. The dataset is then further split per class ($P(X \, | \, Y \in \{0, 1\})$). $\theta$ is the model trained on that dataset. In the second row, we show the distributions that would be produced ($P(X \, | \, \hat{Y} \in \{0, 1\})$) if $\theta$ were used to determine the intervention where $\hat{Y} \in \{0, 1\}$. As shown, the two distributions (Group A and B) are very different, and thus cannot be used to determine if the label / outcome distributions ($P(Y)$) are significantly different. In the third row, we show the distributions that would be produced ($P(X \, | \, \hat{Y} \in \{0, 1\})$) if intervention testing (randomized predictions) were used to determine the intervention. As shown, the two distributions are nearly identical and thus can be used to determine if the outcome / label distributions ($P(Y)$) are significantly different (and therefore Outcome Performative).}
    \label{fig:OPAB_motivation}
\end{figure}

Recall that OP-AB uses intervention testing to detect Outcome Performativity. The motivations for this decision are described using Figure \ref{fig:OPAB_motivation}. Naively, it might be tempting to simply compare whether the outcome distributions conditioned on a classifier's predictions $\hat{y}$ are dissimilar:

\begin{equation}\label{eq:HYPOTHESIS_TEST_MOTIV}
    H(P(Y \, | \, \hat{y} = 0), P(Y \, | \, \hat{y} = 1)) \leq \delta
\end{equation}

Here, $H$ is a function that calculates the dissimilarity of the two distributions (e.g. a hypothesis test)
\footnote{If a distance metric is used, the equation will then instead be: $H(P(Y \, | \, \hat{y} = 0), P(Y \, | \, \hat{y} = 1)) \geq \delta$.} 
and $\delta$ is the threshold that needs to be reached in order to determine if that dissimilarity is significant. 
However, this is \emph{not} sufficient, as these results are only meaningful if the instances that make up the two groups come from the same distribution (i.e. $P(X \, | \, \hat{y}= 0) \approx P(X \, | \, \hat{y}= 1)$). In contradiction, the very aim of fitting a classifier is to maximize the dissimilarity between $P(X \, | \, \hat{y}= 0)$  and $P(X \, | \, \hat{y}= 1)$. This is shown in the second row of Figure \ref{fig:OPAB_motivation}. Using these predictions will often falsely detect Outcome Performativity.
To combat this phenomena, OP-AB takes inspiration from Randomized Controlled Trials by using random sampling to assign predictions, ensuring $P(X \, | \, \hat{y}= 0) \approx P(X \, | \, \hat{y}= 1)$, removing the effects of confounding variables, and preventing false detections in settings with no Outcome Performativity. This is shown in row 3 of Figure \ref{fig:OPAB_motivation}.

\section{Proofs} \label{sect:proofs}

In this section we provide proofs for Equations \ref{eq:N_APPROX_SIMPLE}, \ref{eq:N_APPROX_FEATURE}, and \ref{eq:N_APPROX_ADAM}. All proofs start with same assumption about estimating the number of samples needed per intervention group $N$: \\

\textit{Assumption:} The task is a binary classification task such that an outcome $y \in Y = \{0, 1\}$. \\
\textit{Assumption:} Predictions $\hat{y} \in \widehat{Y} = \{0, 1\}$ are assigned to instances $x \in X$ at random. \\
\textit{Assumption:} A Chi-Squared Test will be used to determine if two outcome distributions are significantly dissimilar.

If these assumptions hold, the $\chi^2$ value of the calculated using the following Equation:

\begin{equation}\label{eq:CHI_SQUARED}
    \chi^2 = \sum\frac{(O_i - E_i)^2}{E_i}
\end{equation}

where $E_i$ is the expected outcome and $O_i$ is the observed outcome of a prediction-outcome pair $i = (y, \hat{y})$. These values are calculated by creating a contingency / frequency table from the prediction-outcome pairs that are observed.

\subsection{Simple Model of Outcome Performativity}

In the Simple Model of Outcome Performativity (Sect. \ref{sect:SOP}), outcomes are determined by:

\begin{equation}\label{eq:SIMPLE_MODEL}
    p* = \begin{cases}
        \alpha_0 & \verb|if | \hat{y} = 0 \\
        \alpha_1  &\verb|if | \hat{y} = 1 \\
    \end{cases}
\end{equation}

where $\alpha_c = P(y=1 \, | \, \hat{y} = c)$ is the probability that the outcome is $y=1$ for when the prediction is $\hat{y} = c \in \{0, 1\}$ (binary classification task). Using Equation \ref{eq:SIMPLE_MODEL}, we can populate a contingency table that would be generated (on average) for a given number of samples per intervention group $N$ and $\alpha_0, \alpha_1 \in [0.0, 1.0]$ as shown in Table \ref{tab:SIMPLE_MODEL_OBSERVED}

\begin{table}[ht]
    \centering
    \begin{tabular}{c|c|c|c}
         & \textbf{$\hat{y} = 0$} & \textbf{$\hat{y} = 1$} & \textbf{Totals} \\
    \hline
         \textbf{$y = 0 $} & $(1-\alpha_0)N$ & $(1-\alpha_1)N$ & $N(2 - \alpha_0 - \alpha_1)$ \\
    \hline
         \textbf{$y = 1 $} & $\alpha_0N$ & $\alpha_1N$ & $N(\alpha_0 + \alpha_1)$ \\
    \hline
         \textbf{Totals} & $N$ & $N$ & $2N$ \\
    \end{tabular}
    \caption{Average observed Contingency Table in the Simple Model of Outcome Performativity given $N$, $\alpha_0$ and $\alpha_1$.}
    \label{tab:SIMPLE_MODEL_OBSERVED}
\end{table}

We can then calculate the Expected Contingency Table that would occur if the Null Hypothesis of the Chi-Squared Test was true as shown in Table \ref{tab:SIMPLE_MODEL_EXPECTED}.

\begin{table}[ht]
    \centering
    \begin{tabular}{c|c|c}
         & \textbf{$\hat{y} = 0$} & \textbf{$\hat{y} = 1$} \\
    \hline
         \textbf{$y = 0 $} & $\frac{N}{2}(2 - \alpha_0 - \alpha_1)$ & $\frac{N}{2}(2 - \alpha_0 - \alpha_1)$ \\
    \hline
         \textbf{$y = 1 $} & $\frac{N}{2}(\alpha_0 + \alpha_1)$ & $\frac{N}{2}(\alpha_0 + \alpha_1)$ \\
    \end{tabular}
    \caption{Expected Contingency Table in the Simple Model of Outcome Performativity if the Null Hypothesis for the Chi-Squared Test is true for a given $N$, $\alpha_0$ and $\alpha_1$.}
    \label{tab:SIMPLE_MODEL_EXPECTED}
\end{table}

We then plug the values from Tables \ref{tab:SIMPLE_MODEL_OBSERVED} and \ref{tab:SIMPLE_MODEL_EXPECTED} into Equation \ref{eq:CHI_SQUARED} which will give us:

\begin{equation}
     \chi^2 = 2N\frac{(\alpha_0 - \alpha_1)^2}{2(\alpha_0 + \alpha_1) - (\alpha_0 + \alpha_1)^2}
\end{equation}

which we can rearrange to estimate $N$:

\begin{equation} \label{eq:N_APPROX_SIMPLE_APP}
    N = \chi^2\frac{ 2(\alpha_0 + \alpha_1) - (\alpha_0 + \alpha_1)^2}{2(\alpha_0 - \alpha_1)^2}
\end{equation}

\begin{figure}[tb]
\centering
\includegraphics[width=0.8\columnwidth]{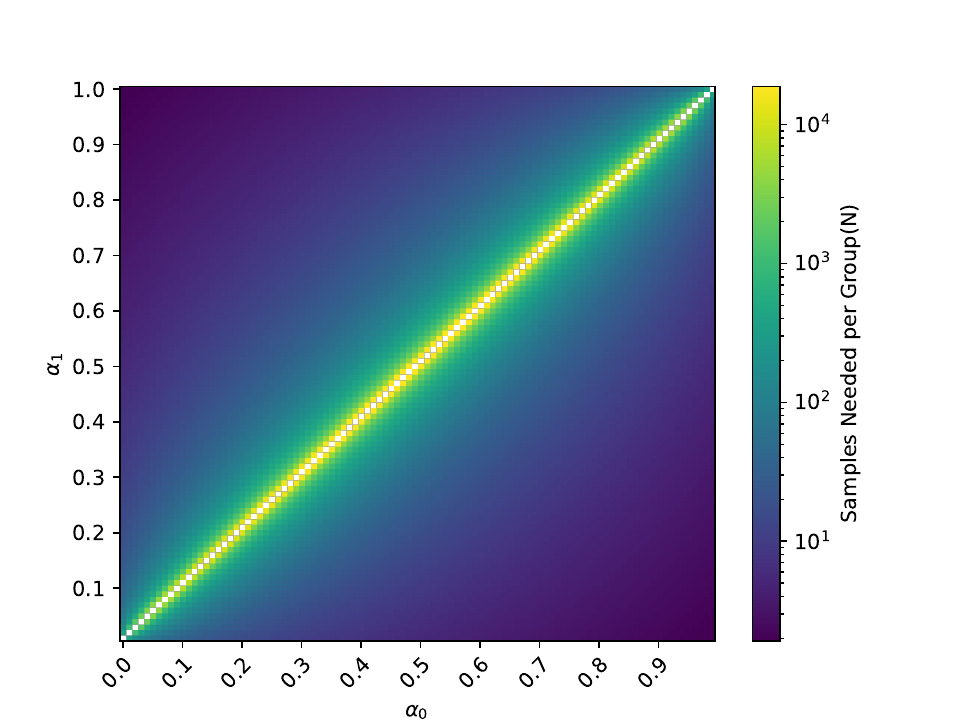}
\caption{The estimated number of instances needed per group ($N$) to detect Outcome Performativity (Equation \ref{eq:N_APPROX_SIMPLE}) across varying $\alpha_0$ and $\alpha_1$ values. The two main findings presented in this figure are that Outcome Performativity is undetectable when $\alpha_0 \approx \alpha_1$ and only a few samples are needed to detect Outcome Performativity when $|\alpha_0 - \alpha_1| \to 1.0$.}
\label{fig:N_THRESHOLD_SIMPLE}
\end{figure}

Here $\chi^2$ is the Chi-Squared statistic that you'd want to achieve. In the work, we set $\chi^2 = 3.841$ which is the threshold required to get a significant result for $p=0.05$ with one degree of freedom. Figure \ref{fig:N_THRESHOLD_SIMPLE} shows visualization of Equation \ref{eq:N_APPROX_SIMPLE_APP}.

\subsection{Model-based Outcome Performativity}

The model-based model \cite{MendlerDunnerDingWang2022} takes into account an instance's position in the feature space. Given a set of labelled instances $(X, \bar{Y})$, a linear model $f_{\theta} \to \{0, 1\}$ is trained and its weights $\theta$ observed. Using $\theta$, we simulate Outcome Performativity using:

\begin{equation} \label{eq:feature_dependent_model}
    y = f_{\theta}(x + \beta_{\hat{y}}.\theta)
\end{equation}

where $\beta_{\hat{y}}$ is the strength of the Outcome Performativity for prediction $\hat{y} = c$. If $\beta_c < 0.0$, it biases instances towards an outcome of $y = 0$. If $\beta_c > 0.0$, it biases instances towards an outcome of $y = 1$. If $\beta_c = 0.0$, predictions are non-Outcome Performative.
Using Equation \ref{eq:feature_dependent_model}, we can populate a contingency table that would be generated (on average) for a given number of samples per intervention group $N$ and $\beta_0, \beta_1 \in [-1.0 1.0]$ as shown in Table \ref{tab:FEATURE_MODEL_OBSERVED}

\begin{table}[ht]
    \centering
    \begin{tabular}{c|c|c|c}
         & \textbf{$\hat{y} = 0$} & \textbf{$\hat{y} = 1$} & \textbf{Totals} \\
    \hline
         \textbf{$y = 0 $} & $(1-p_{\beta0})N$ & $(1-p_{\beta1})N$ & $N(2 - p_{\beta0} - p_{\beta1})$ \\
    \hline
         \textbf{$y = 1 $} & $p_{\beta0}N$ & $p_{\beta1}N$ & $N(p_{\beta0} + p_{\beta1})$ \\
    \hline
         \textbf{Totals} & $N$ & $N$ & $2N$ \\
    \end{tabular}
    \caption{Average observed Contingency Table in the Model-based assumption class of Outcome Performativity given $N$, $\beta_0$ and $\beta_1$.}
    \label{tab:FEATURE_MODEL_OBSERVED}
\end{table}

In Table \ref{tab:FEATURE_MODEL_OBSERVED}, the number of instances that would have an outcome of $y=1$ given a prediction of class $\hat{y} = c$ when using model $f_\theta$ is given by $p_{\beta c} = P(Y=1 \, | \, \widehat{Y} = c \, , \, \theta)$. This contingency table has the same form as Table \ref{tab:SIMPLE_MODEL_OBSERVED}. The derivation for which we know is:

\begin{equation} \label{eq:N_APPROX_FEATURE_APP}
    N = \chi^2\frac{ 2(p_{\beta0} + p_{\beta 1}) - (p_{\beta 0} + p_{\beta 1})^2}{2(p_{\beta 0} - p_{\beta 1})^2}
\end{equation}

\begin{figure}[tb]
\centering
\includegraphics[width=0.8\columnwidth]{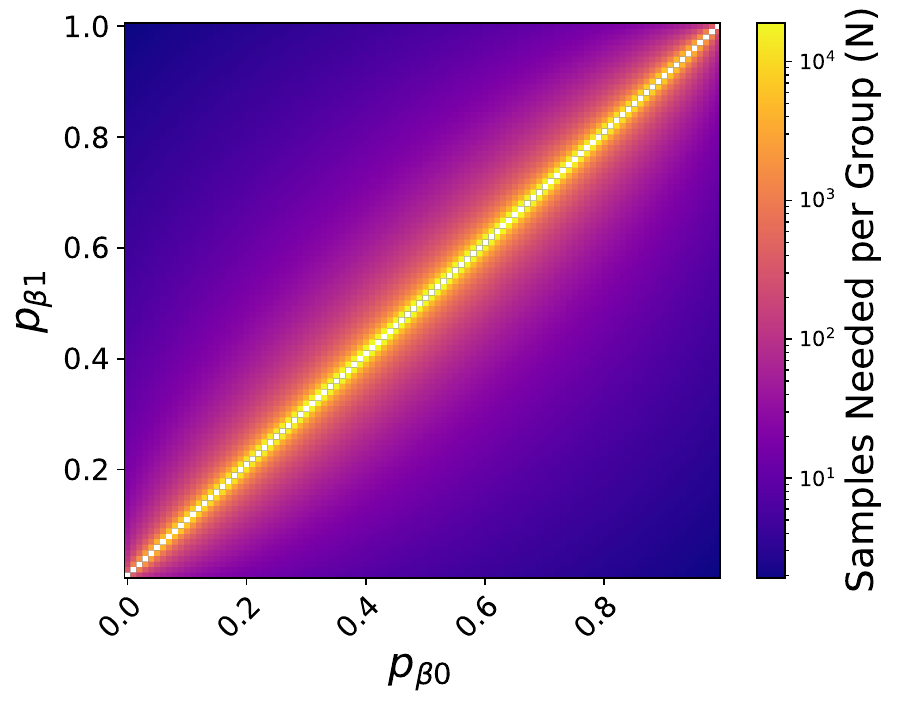}
\caption{The estimated number of instances needed per group ($N$) to detect Outcome Performativity (Equation \ref{eq:N_APPROX_FEATURE_APP}) across varying $p_{\beta0}$ and $p_{\beta1}$ values. This figure is shows that model-based Outcome Performativity behaves similarly to Simple Model Outcome Performativity.}
\label{fig:N_THRESHOLD_FEATURE}
\end{figure}

The visualization (which looks the same as Figure \ref{fig:N_THRESHOLD_SIMPLE}) is shown in Figure \ref{fig:N_THRESHOLD_FEATURE} where the two main findings, as is the case in the Simple Model of Outcome Performativity presented are that Outcome Performativity is undetectable when $p_{\beta0} \approx p_{\beta1}$ and only a few samples are needed to detect Outcome Performativity when $|p_{\beta0} - p_{\beta1}| \to 1.0$. Of course, this derivation is not necessarily useful unless you know $p_{\beta0}$ and $p_{\beta1}$. This can be done for any arbitrary set of features $X$ and training labels $Y'$ where $f_\theta$ trains on $X$, and $Y'$, and then, for the desired $\beta_0$ and $\beta_1$ pair, you calculate $p_{\beta0}$ and $p_{\beta1}$.

\subsection{Misclassification-based Outcome Performativity}

Recall the definition of misclassification-based Outcome Performativity: Given a set of $T$ instances and desirable outcomes $(X, \overline{Y})$, the outcome $y \in Y = \{0, 1\}$ of an instance $x \in X$, given desirable outcome $\bar{y} \in \overline{Y} = \{0, 1\}$ and prediction $\hat{y} \in \widehat{Y} = \{0, 1\}$ is defined as:

\begin{equation}
    p^* = \begin{cases}
        1.0 & \verb|if | \hat{y} = 1 \verb| and | \bar{y} = 1 \\
        0.0 & \verb|if | \hat{y} = 0 \verb| and | \bar{y} = 0 \\
        \gamma_1 & \verb|if | \hat{y} \neq 1 \verb| and | \bar{y} = 1 \\
        1 - \gamma_0 & \verb|if | \hat{y} \neq 0 \verb| and | \bar{y} = 0 \\
    \end{cases}
\end{equation}

where $\lambda_c$ is the likelihood that a desired outcome is realized despite being misclassified:

\begin{equation}
    \lambda_c = P(y=c \, | \, \bar{y} = c  \verb| and | \hat{y} \neq c)
\end{equation}

Applying the same process as before, we can populate a contingency table that would be generated (on average) for a given number of samples per intervention group $N$ and $\gamma_0, \gamma_1 \in [0.0, 1.0]$ as shown in Table \ref{tab:ADAM_MODEL_OBSERVED}

\begin{table}[ht]
    \centering
    \begin{tabular}{c|c|c|c}
         & \textbf{$\hat{y} = 0$} & \textbf{$\hat{y} = 1$} & \textbf{Totals} \\
    \hline
         \textbf{$y = 0 $} & $N(\gamma_1 p_0 - \gamma_1 + 1)$ & $N \gamma_0p_0$ & $N(\gamma_0 p_0 + \gamma_1p_0 - \gamma_1 + 1)$ \\
    \hline
         \textbf{$y = 1 $} & $N \gamma_1 (1 - p_0)$ & $N (-\gamma_0 p_0 + 1)$ & $N(-\gamma_0p_0 -\gamma_1(p_0 - 1) + 1)$\\
    \hline
         \textbf{Totals} & $N$ & $N$ & $2N$ \\
    \end{tabular}
    \caption{Average observed Contingency Table in misclassification-based Outcome Performativity given $N$, $p0$, $\gamma_0$ and $\gamma_1$.}
    \label{tab:ADAM_MODEL_OBSERVED}
\end{table}

Note that in this model we have to also include $p_0$ which describes the likelihood that a randomly drawn sample $x \in X$ will have a desirable outcome $\bar{y} = 0$. We can then calculate the Expected Contingency Table that would occur if the Null Hypothesis of the Chi-Squared Test was true as shown in Table \ref{tab:ADAM_MODEL_EXPECTED}.

\begin{table}[ht]
    \centering
    \begin{tabular}{c|c|c}
         & \textbf{$\hat{y} = 0$} & \textbf{$\hat{y} = 1$} \\
    \hline
         \textbf{$y = 0 $} & $\frac{N}{2}(\gamma_0p_0 + \gamma_1p_0 -\gamma_1 + 1)$ & $\frac{N}{2}(\gamma_0p_0 + \gamma_1p_0 -\gamma_1 + 1)$ \\
    \hline
         \textbf{$y = 1 $} & $\frac{N}{2}(-\gamma_0p_0 - \gamma_1p_0 + \gamma_1 + 1)$ & $\frac{N}{2}(-\gamma_0p_0 - \gamma_1p_0 + \gamma_1 + 1)$ \\
    \end{tabular}
    \caption{Expected Contingency Table in a misclassification-based Outcome Performative settings if the Null Hypothesis for the Chi-Squared Test is true for a given $N$, $p_0$, $\gamma_0$ and $\gamma_1$.}
    \label{tab:ADAM_MODEL_EXPECTED}
\end{table}

We then plug the values from Tables \ref{tab:ADAM_MODEL_OBSERVED} and \ref{tab:ADAM_MODEL_EXPECTED} into Equation \ref{eq:CHI_SQUARED} which will give us:

\begin{equation}
     \chi^2 = -\frac{2N \times (\gamma_0p_0 -\gamma_1p_0 + \gamma_1 - 1)^2}{(\gamma_0p_0 +\gamma_1p_0 - \gamma_1 - 1)(\gamma_0p_0 +\gamma_1p_0 - \gamma_1 + 1)}
\end{equation}

which we can rearrange to estimate $N$:

\begin{equation}\label{eq:N_APPROX_ADAM_APP}
    N = -\frac{\chi^2\times(\gamma_0p_0 +\gamma_1p_0 - \gamma_1 - 1)(\gamma_0p_0 +\gamma_1p_0 - \gamma_1 + 1)}{2 \times (\gamma_0p_0 -\gamma_1p_0 + \gamma_1 - 1)^2}
\end{equation}

\begin{figure}[tb]
\centering
\includegraphics[width=0.8\columnwidth]{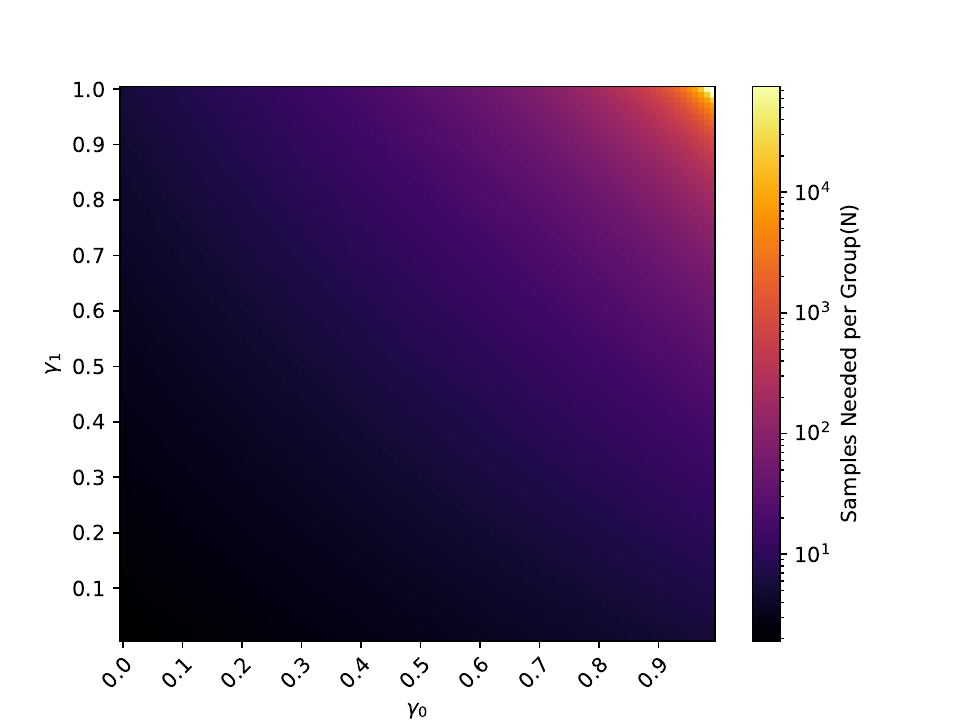}
\caption{The estimated number of instances needed per group ($N$) to detect Outcome Performativity (Equation \ref{eq:N_APPROX_ADAM}, $p_0 = 0.5$) across varying $\gamma_0$ and $\gamma_1$ values. This figure shows that it is increasingly difficult to detect Misclassification-based Outcome Performativity as $\gamma_0$ and $\gamma_1 \to 1.0$.}
\label{fig:N_THRESHOLD_ADAM}
\end{figure}

Note that getting to Equation \ref{eq:N_APPROX_ADAM} required quite a few steps which we had to validate using \textit{Sympy}. Figure \ref{fig:N_THRESHOLD_ADAM} shows visualization of Equation \ref{eq:N_APPROX_ADAM}.

\subsubsection{Visualizations of Varying Initial Label Distributions}

\begin{figure}[htb]
    \centering
    \includegraphics[width=\linewidth]{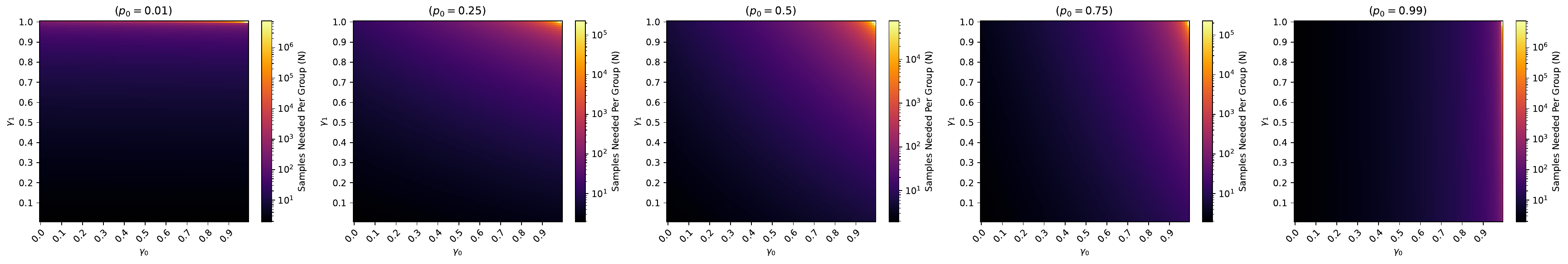}
    \caption{Visualizations of Equation \ref{eq:N_APPROX_ADAM} across varying values of $p_0$. The values on these heatmaps report the estimated number of samples needed per intervention group ($N$) to detect Outcome Performativity when $\chi^2 = 3.841$.}
    \label{fig:P0_VIZ}
\end{figure}

Given that Equation \ref{eq:N_APPROX_ADAM} requires that $p_0$ be stipulated to know the estimated number of samples per intervention needed $N$ to detect Outcome Performativity, we include plots for various values $p_0$ as shown in Figure \ref{fig:P0_VIZ}. Here we have assumed a $\chi^2 = 3.841$. The plots clearly show that as $p_0 \to 0.0$, the number of instances $N$ required per intervention increases exponentially predominantly dependent on the misclassification response of class 0 ($\gamma_0)$. Conversely, as $p_0 \to 1.0$, the number of instances $N$ required per intervention increases exponentially predominantly dependent on the misclassification response of class 1 ($\gamma_1)$. These results make sense given that the detecting Outcome Performativity in these settings becomes increasingly dependent on the prediction response of the minority class. This is also supported by the two orders of magnitude more instances that are required to detect subtle ($\gamma_0$, $\gamma_1$ close to 1.0) Outcome Performativity.

\section{Experimental Design}\label{sect:experimental_design}

In this work, we made of the following datasets for our misclassification and model-based assumption classes of Outcome Performativity. Note that in order to get the datasets to work with the model-based Outcome Performativity Model, we could only make use of numerical features which are listed in the description below:

\begin{enumerate}
    \item \verb|breast| \verb|cancer| \cite{DATASET_BREAST_CANCER}: The Breast Cancer Wisconsin (Diagnostic) Data Set. It consists of 569 instances of which 357 are benign and 212 malignant. We use all 30 features.
    
    \item \verb|diabetes| \cite{DATASET_DIABETES}: The Pima Indians Diabetes Database. It consists of 768 instances of which 268 belong to Class 1 (The others Class 0). We use all 8 features.
    
    \item \verb|adult| \verb|census| \cite{DATASET_ADULT_CENSUS}: The 1994 Adult Census Income dataset. It has $\approx 32\, 000$ instances of which $76\%$ are of Class $\leq 50k$ which the others are of class $>50k$. We make use of the \verb|age|, \verb|fnlwgt|, \verb|education.num|, \verb|capital.gain|, \verb|capital.loss|, and \verb|hours.per.week| features.
    
    \item \verb|kickstarter| \cite{DATASET_KICKSTARTER}: The Funding Successful Kickstarter Projects dataset. It contains $108\,129$ instances of which $73\, 568$ are Class $0$. The others are Class $1$. We make use of the \verb|goal| and \verb|backers\_count| features.
    
    \item \verb|titanic| \cite{DATASET_TITANIC}: The Titanic Survival Prediction dataset. It consists of $891$ instances of which $549$ are Class $0$ while the others are Class 1. We make use of the \verb|Age|, \verb|SibSp|, \verb|Parch|, and \verb|Fare| features.
    
    \item \verb|loan| \cite{DATASET_LOAN}: A Loan Approval Dataset. It consists of 4269 instances of which $62\%$ are labelled \textit{Approved}. The others are labelled \textit{Rejected}. We use all 9 numerical features.
\end{enumerate}

For all of the Experiments conducted in sections \ref{sect:ADAM_PERF} and \ref{sect:MBOP}, each parameter set was evaluated over $100$ replicates. The hypothesis test $H$ given to OP-AB was the Chi-Squared test. In the case where the outcome contingency table contained a $0$, Fisher's Exact test was used instead. A successful detection meant that the $p$ value returned by the hypothesis test was $< \delta = 0.05$. A pseudorandom number generator was used to ensure reproducibility. All experiments were run in Google Colab using the Python 3 Google Compute Engine Backend.

\subsection{Open Bandits Dataset}

For our real-world case study, we made use of the Open Bandits Dataset (OBD \cite{SaitoShunsukeMegumi2020}). OBD was constructed using multi-armed bandit policies on the fashion e-commerce platform ZOZOTOWN for the off-policy evaluation of recommender systems. Each instance represents a user impression containing feature values, item recommendations and their placement (clothing items to choose from and where they appear on the store webpage), and click indicators (was the recommended clothing item clicked on) as an outcome. For a subset of the data, the placement of each recommended item on the store page (left, centre, or right) was determined using random sampling. We use this subset and repurpose the placement of item recommendation as predictions and click indications as outcomes.

For preprocessing, we take all of the randomly sampled impressions and extract the \verb|position| and \verb|click| values. The former is our set of predictions $\hat{Y}$ and the latter the set of outcomes $Y$. To ensure that both variables are binary, we filter out any recommendations with \verb|position == 2| which corresponds to $centre$ on the webpage. The rest of the preprocessing proceeds as described in the Sect. \ref{sect:EXPERIMENTAL_DESIGN}, we first test to see if Outcome Performativity is present in each sub-dataset (\verb|men|, \verb|women|). We then perform random subsampling procedures over $1000$ replicates to evaluate OP-AB.

\section{Additional Results}

In this section we report on the results for the \verb|Titanic| and \verb|Loan| datasets for both the misclassification and feature-based Models of Outcome Performativity. Additionally, we report on the additional experiments that we conducted. Namely: The efficacy of OP-AB when using different statistical tests, relationship between sample complexity $N$ and $\delta$, and demonstrating the importance $N$ plays in determining the detection capabilities of OP-AB. The section concludes with a comparison of OP-AB and the Online Adam detection method described in \cite{AdamChangHaibe2020}.

\subsection{Misclassification-based Outcome Performativity}

\begin{figure}[tb]
    \centering
    \includegraphics[width=\linewidth]{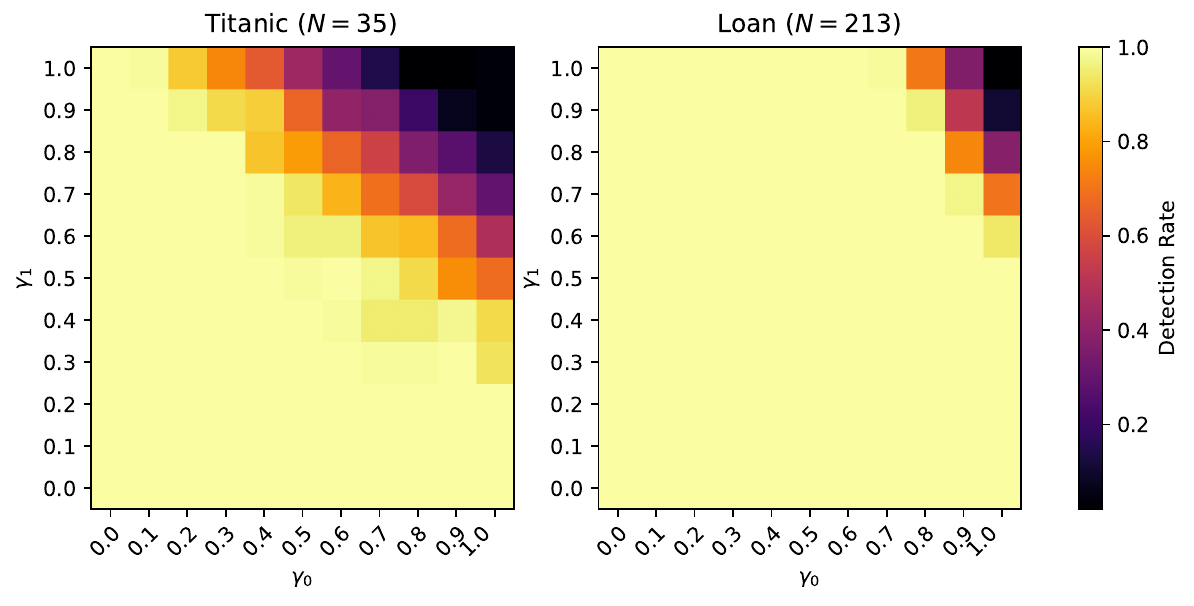}
    \caption{Results of applying OP-AB to detect misclassification-based Outcome Performativity. These results are in keeping with our main findings. OP-AB exhibits a low false detection rate (i.e. $\gamma_0 = \gamma_1 = 1.0$) and if $N$ is larger, OP-AB can detect subtler Outcome Performative Effects.}
    \label{fig:ADAM_MODEL_SUPP}
\end{figure}

Figure \ref{fig:ADAM_MODEL_SUPP} reports on the results of applying OP-AB to detect misclassification-based Outcome Performativity on the \verb|Titanic| and \verb|Loan| datasets. These results are in keeping with our main findings. OP-AB exhibits a low false detection rate (i.e. $\gamma_0 = \gamma_1 = 1.0$). If $N$ is larger, OP-AB can detect subtler Outcome Performative Effects.

\subsection{Model-based Outcome Performativity}

\begin{figure}[htb]
    \centering
    \includegraphics[width=\linewidth]{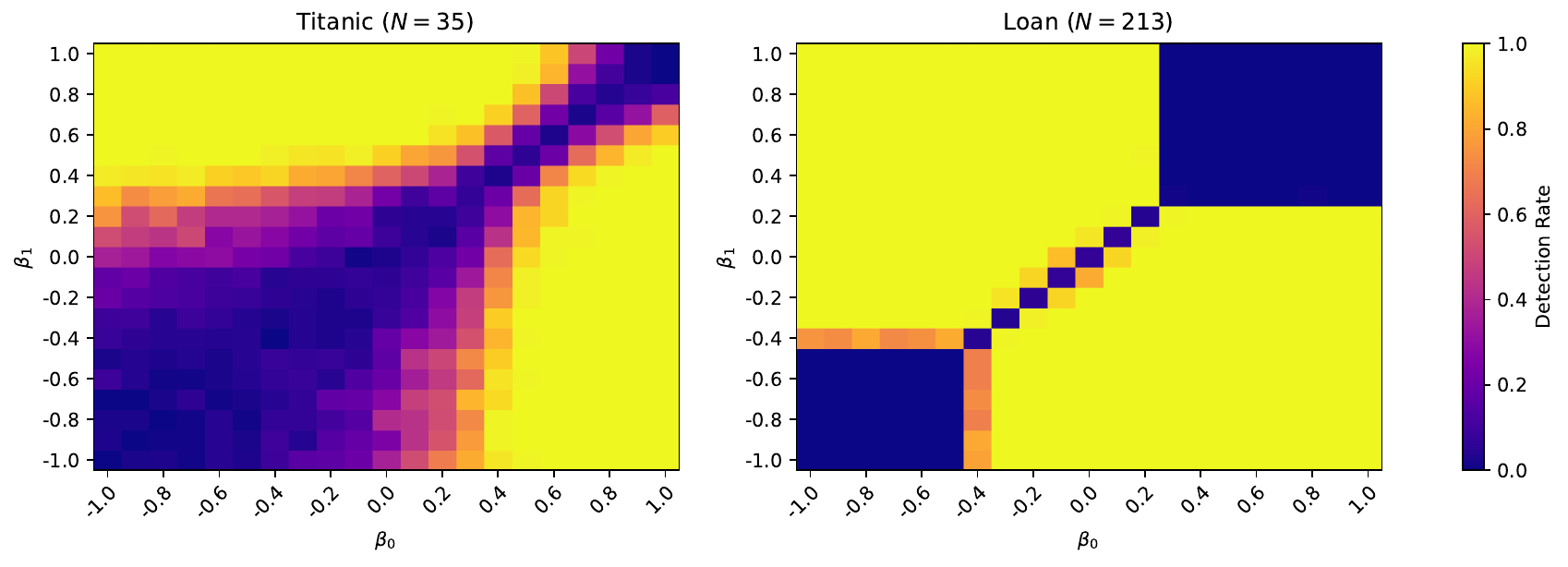}
    \caption{Results of applying OP-AB to detect model-based Outcome Performativity. These results are in keeping with our main findings. OP-AB exhibits a low false detection rate when $\beta_0 = \beta_1$) and if $N$ is larger, OP-AB can detect subtler Outcome Performative Effects.}
    \label{fig:FEATURE_DEPENDENT_MODEL_SUPP}
\end{figure}

Figure \ref{fig:FEATURE_DEPENDENT_MODEL_SUPP} reports on the results of applying OP-AB to detect model-based Outcome Performativity on the \verb|Titanic| and \verb|Loan| datasets. These results are in keeping with our main findings. OP-AB exhibits a low false detection rate when $\beta_0 = \beta_1$) because the outcome distributions per intervention are identical and thus non-Outcome Performative. If $N$ is larger, OP-AB can detect subtler Outcome Performative Effects. Figure \ref{fig:FEATURE_DEPENDENT_MODEL_SUPP} also shows the limitation of the model-based assumption class of Outcome Performativity as features with extremely skewed Gaussian or Bimodal distributions (as in the case of the \verb|Loan| dataset) can cause significant portions of the features to change outcomes at small  $\beta_n$ values. This is not a limitation of OP-AB, but rather of the assumption class of Outcome Performativity.

\subsection{Using OP-AB with Other Statistical Tests}\label{sect:additional_hypothesis_tests}

\begin{figure}[htb]
    \centering
    \includegraphics[width=\linewidth]{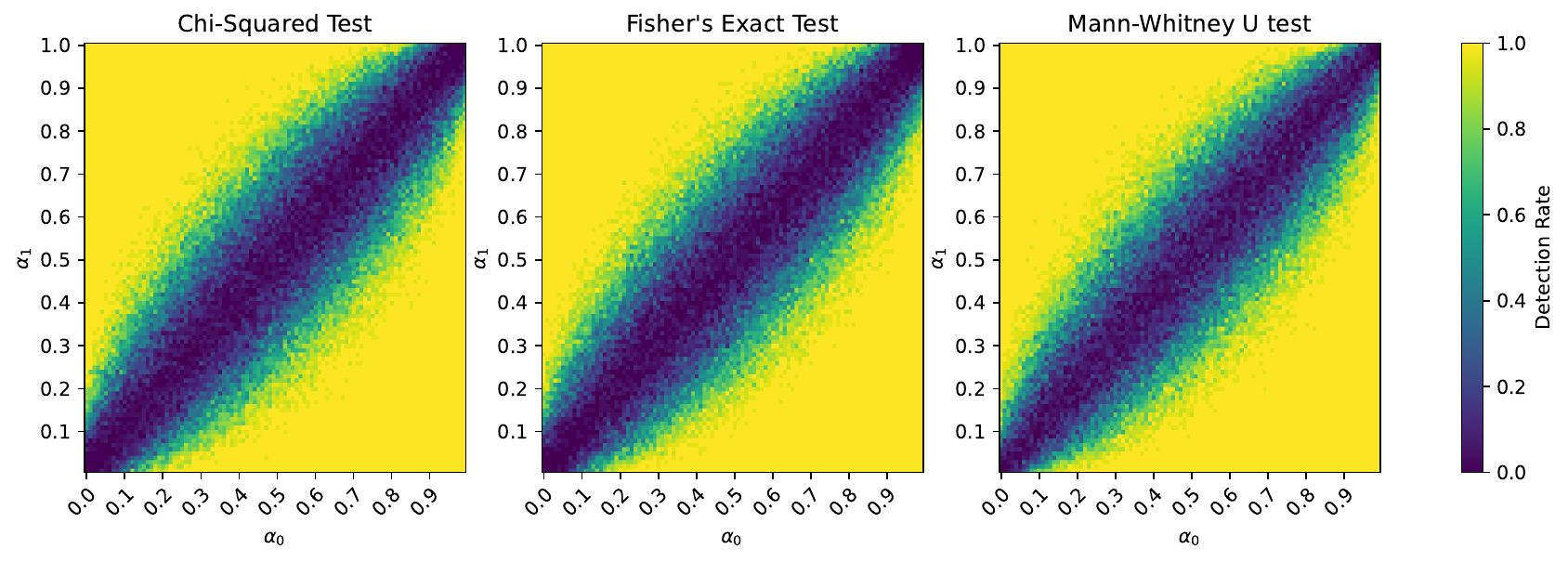}
    \caption{Results of using various Hypothesis Tests with OP-AB to detect Outcome Performativity. These results show that OP-AB is agnostic to the $H$ used to detect Outcome Performativity (as long as $H$ is appropriately applied).}
    \label{fig:STATS_TESTS}
\end{figure}

In this work we primarily used the Chi-Squared Test to determine if the outcome distributions per intervention were significant. OP-AB is hypothesis test agnostic so we wanted to test the efficacy of OP-AB over different Statistical Tests. Figure \ref{fig:STATS_TESTS} reports the detection rate of OP-AB applied to the Simple Model of Outcome Performativity using the Chi-Squared, Fisher's Exact of Mann-Whitney U tests. Each cell is averaged over $100$ replicates. We conducted these experiments with $N=50$. The results reported in Figure \ref{fig:STATS_TESTS} clearly show that across the statistical tests evaluated, the detection rate of OP-AB remains approximately the same. The Mann-Whitney U test seems slightly more unreliable when the Outcome Performativity response is low (i.e. small $|\alpha_0 - \alpha_1|$. These results reveal two additional insights: First, $N$ is without a doubt the most important parameter in OP-AB. If too few instances are used in the intervention testing, OP-AB will not detect Outcome Performativity. Second, the trends estimated by Eq. \ref{eq:N_APPROX_SIMPLE}, \ref{eq:N_APPROX_ADAM} and \ref{eq:N_APPROX_FEATURE} in the main paper capture the statistical test-agnostic dynamics of the Outcome Performativity Models in this work.

\subsection{Varying Detection Threshold $\delta$}

\begin{figure}
     \centering
     \begin{subfigure}[b]{\linewidth}
         \centering
         \includegraphics[width=\textwidth]{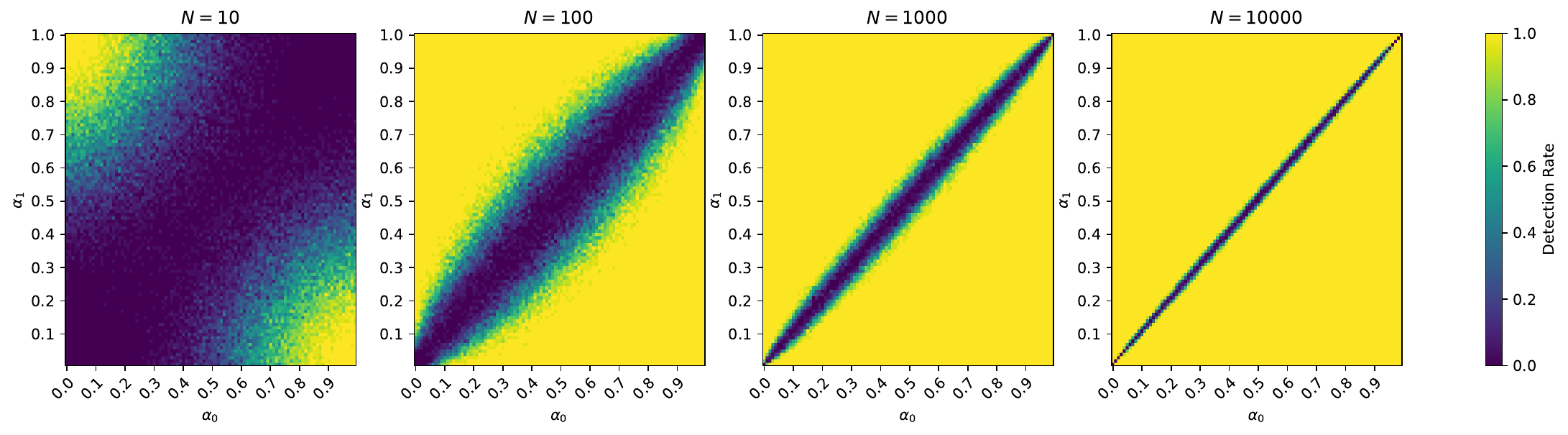}
         \caption{$\delta = 0.01$}
         \label{fig:DELTA_001}
     \end{subfigure}
     \hfill
     \begin{subfigure}[b]{\linewidth}
         \centering
         \includegraphics[width=\textwidth]{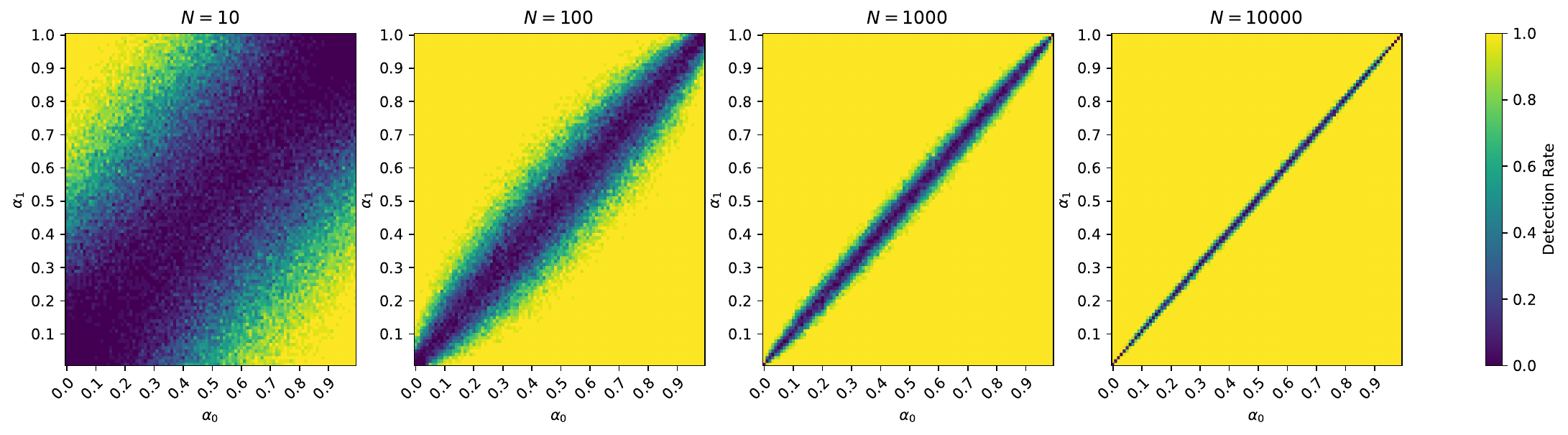}
         \caption{$\delta = 0.05$}
         \label{fig:DELTA_005}
     \end{subfigure}
     \hfill
     \begin{subfigure}[b]{\linewidth}
         \centering
         \includegraphics[width=\textwidth]{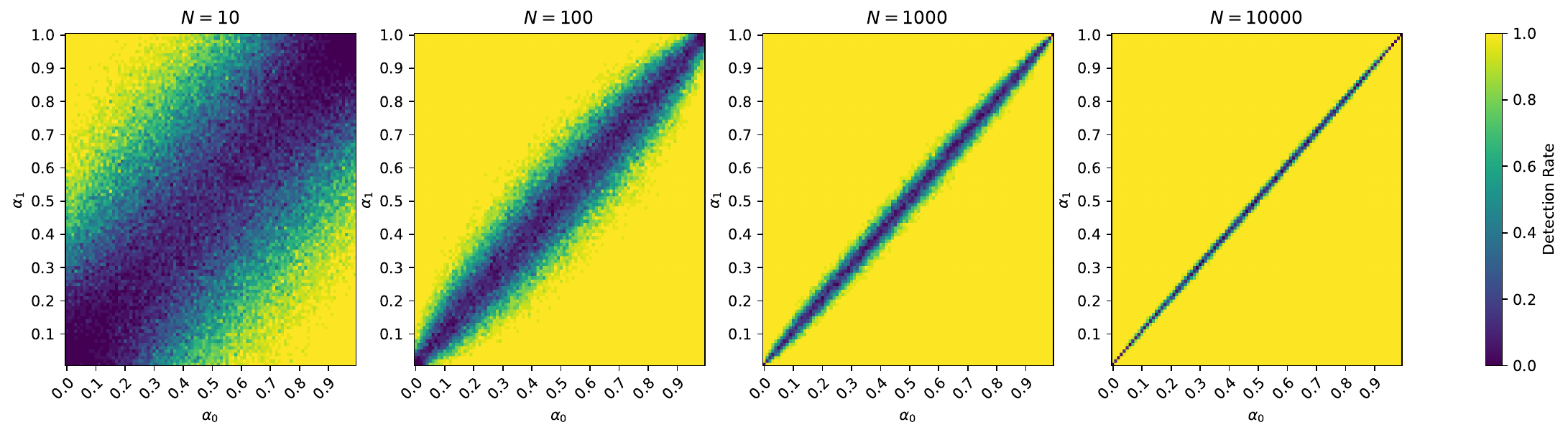}
         \caption{$\delta = 0.1$}
         \label{fig:DELTA_01}
     \end{subfigure}
        \caption{Results from exploring various detection thresholds $\delta$ on the Simple Outcome Performativity Model. These results show that as $\delta$ decreases, sample complexity $N$ increases. The trade-off being that detection events at a smaller $\delta$ is more reliably than a detection event at a larger $\delta$.}
        \label{fig:DELTA_EXPLORATION}
\end{figure}

In OP-AB, the detection threshold $\delta$ corresponds to the necessary p-value required by Hypothesis test $H$ to raise a detection event. Naturally, one would assume that as $\delta$ increases, the OP-AB becomes less restrictive and will more likely flag Outcome Performativity. This should also be doable with fewer random samples. Conversely, as $\delta$ decreases, OP-AB will become more restrictive and less likely to flag Outcome Performativity. A detection event, on overage, will require more samples as $\delta$ decreases. The trade-off when choosing $\delta$ is between sample complexity (number of samples $N$ needed per group) and the reliability of a detection event (a higher $\delta$ will produce more false positives). Figure \ref{fig:DELTA_EXPLORATION} showcases the sample complexity dynamics across various commonly used statistical thresholds. These results confirm our suspicions, as $\delta$ increases, sample complexity $N$ decreases.

\subsection{Increased N on Breast Cancer Dataset} \label{sect:feat_vary_N}

\begin{figure}[htb]
    \centering
    \includegraphics[width=\linewidth]{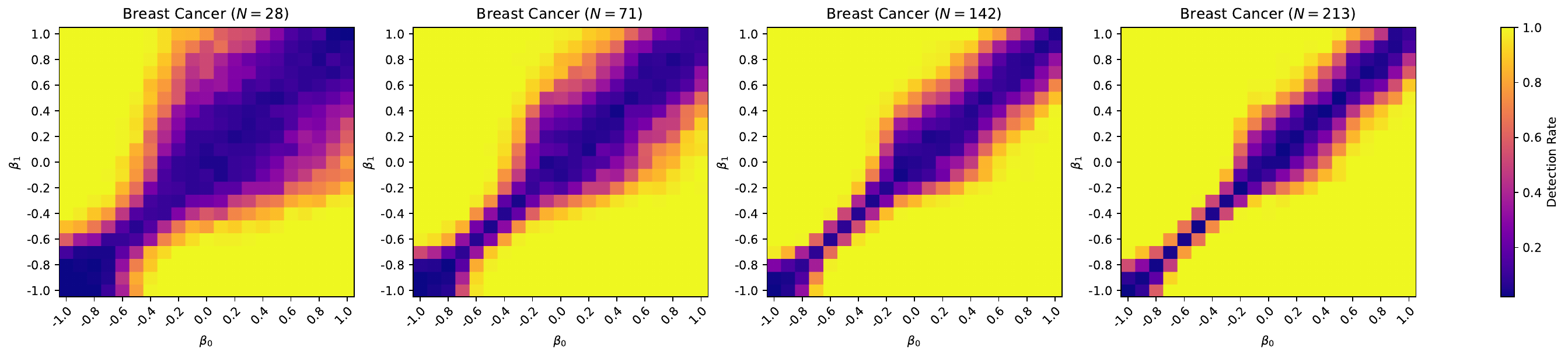}
    \caption{Results demonstrating the importance the number of instances per intervention ($N$) plays on the efficacy of OP-AB. A higher $N$ allows OP-AB to detect Outcome Performativity across a wider range of $\beta_0$ and $\beta_1$ values .}
    \label{fig:N_BREAST_CANCER}
\end{figure}

For the final set of additional experiments, we wanted to empirically demonstrate the importance that $N$ plays in being able to detect Outcome Performativity. Figure \ref{fig:N_BREAST_CANCER} shows these results where we apply OP-AB to the \verb|Breast| \verb|Cancer| dataset imputed with Feature-based Outcome Performativity over an increasing $N$. The results clearly show that as $N$ increases, OP-AB is able to detect Outcome Performativity across a wider range of $\beta_0$ and $\beta_1$ values. Again, we want to note that our results suggest that as the difference in Outcome Performativity responses tend to non-Outcome Performativity ($|\beta_0 - \beta_1| \to 0.0$), the number of instances required per intervention $N \to \infty$.

\subsection{Comparison with Adam Detection}

\begin{figure}[tb]
    \centering
    
    \begin{subfigure}{0.28\textwidth}
        \centering
        \includegraphics[width=\textwidth]{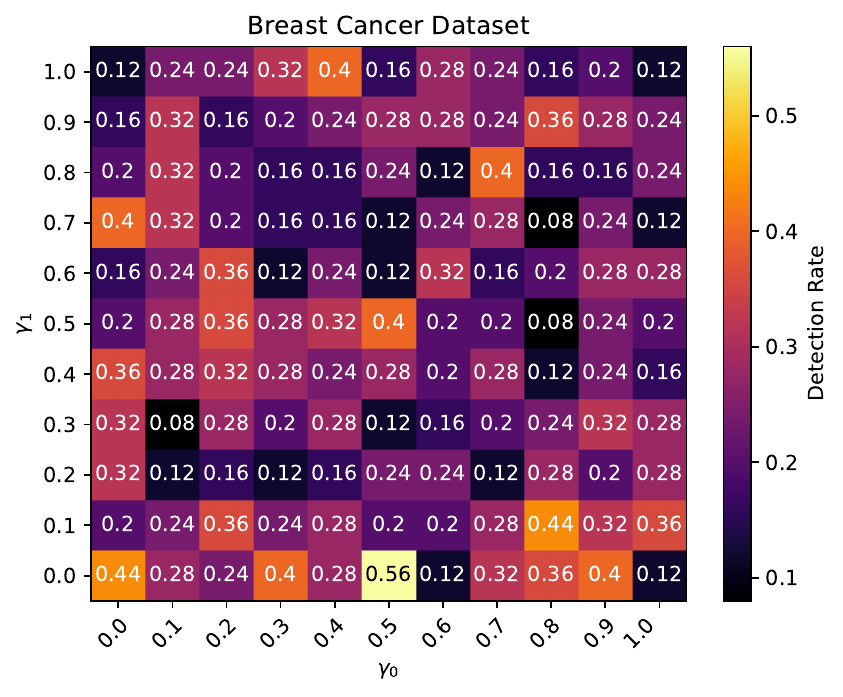}
        \caption{}
        \label{fig:ADAM_DETECTION_BREAST_CANCER}
    \end{subfigure}
    \hfill
    \begin{subfigure}{0.28\textwidth}
        \centering
        \includegraphics[width=\textwidth]{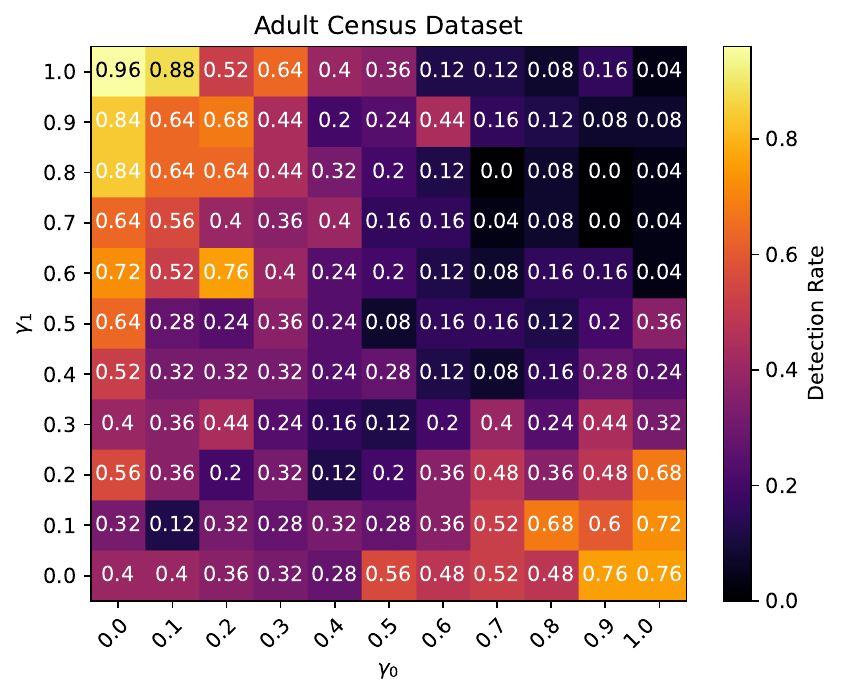}
        \caption{}
        \label{fig:ADAM_DETECTION_ADULT_CENSUS}
    \end{subfigure}
    \hfill
    \begin{subfigure}{0.28\textwidth}
        \centering
        \includegraphics[width=\textwidth]{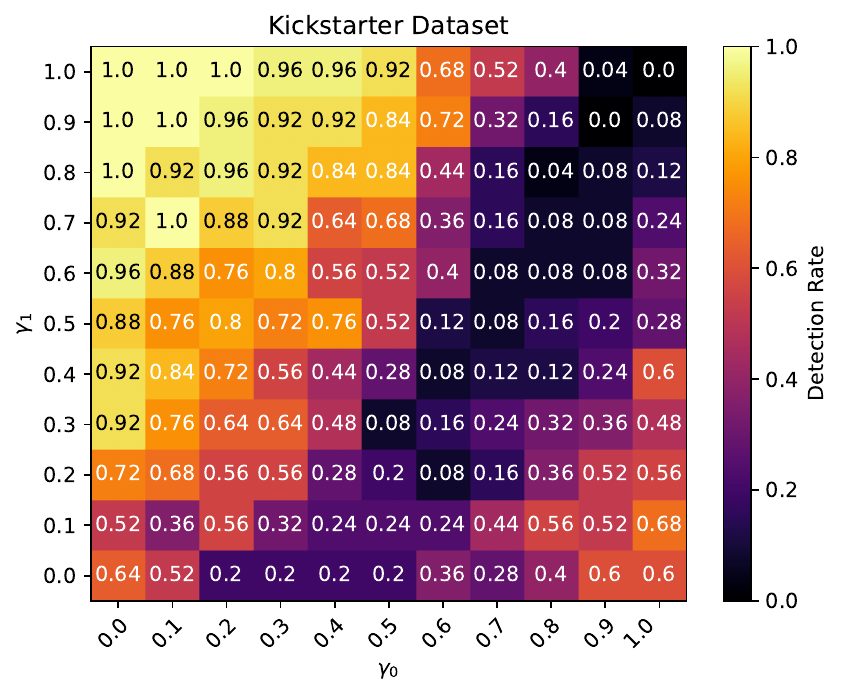}
        \caption{}
        \label{fig:ADAM_DETECTION_Kickstarter}
    \end{subfigure}
    
    \caption{Adam Detection performed on the \textit
{Breast Cancer} (a), \textit{Adult Census} (b) and \textit{Kickstarter} (c) datasets. Overall, the detection rate of Adam Detection is inconsistent. This is partly due to the accuracy of the deployed classifiers which when high, rarely produce misclassifications making detection difficult. The number of instances in each dataset also play a role with more instances increasing the likelihood of detection such as in the \textit{Kickstarter} dataset. Our method, OP-AB, achieves higher and more consistent detection rates with only $10\%$ of the samples used by Adam Detection.}
    \label{fig:ADAM_DETECTION}
\end{figure}

To the best of our knowledge, the only other explicit Outcome Performativity detection method is described in \cite{AdamChangHaibe2020}. In short, this method (which we will call \textit{Adam Detection}) monitors the change in True Positive Rate (TPR) of a deployed model compared to its TPR that it achieved on its historical (training) data. If the change in TPR is determined to be statistically significant, Outcome Performativity is detected. Figures \ref{fig:ADAM_DETECTION} summarizes the results of Adam Detection on several datasets imputed with the Misclassification Outcome Performativity described earlier. In all figures we can see that the detection rates of Adam Detection are lower and more inconsistent when compared to the results obtained by OP-AB. This occurs because Adam Detection requires a deployed model. Under Misclassification Outcome Performativity, the more accurate the deployed model classifier, the harder it is to detect said Outcome Performativity as only misclassified instances are susceptible to having their observed outcomes altered. This is most evident in Figure \ref{fig:ADAM_DETECTION_BREAST_CANCER} where the deployed model achieves high accuracy, rarely misclassifying instances. Adam Detection is also less sample efficient because it requires historical labelled data to initially train a model, and labelled data after the model has been deployed. This is shown in Figures \ref{fig:ADAM_DETECTION_ADULT_CENSUS} and  \ref{fig:ADAM_DETECTION_Kickstarter} where Adam Detection exhibits the same detection rate trends as OP-AB (although less consistent), but it required the whole of the \verb|Adult| \verb|Census| and \verb|Kickstarter| datasets whereas OP-AB achieved better and more consistent detection rates with only $10\%$ of the datasets' instances. Adam Detection is an online algorithm, which is susceptible to both Intrinsic \cite{GamaZliobaiteBifet2014} and non-Outcome Performative \cite{GowerWinterKrempl2025} Concept Drift phenomena.

These results put the primary advantages of OP-AB into perspective. Because OP-AB is an offline algorithm, it is (1) not susceptible to Concept Drift, and (2) it can be used already during the labelling process of a ML model's development lifecycle. This has the added benefit of requiring fewer instances (i.e. better sample efficiency) and it eliminates any potential negative impacts that might occur from a deployed model that is used before the Outcome Performativeness of a problem domain is detected.

\subsubsection{Implementing Adam Detection}

Unfortunately, we were not able to obtain the source code for Adam detection so we had to remake it using the text-only description provided in the referenced paper \cite{AdamChangHaibe2020}. This was slightly problematic as the paper did not describe the Online Updating procedure for the deployed model which made it challenging to know if we replicated Adam detection exactly.

In short, our implementation of Adam will take in an entire static dataset, shuffle it and partition it into $11$ approximately equal sized chunks. The 11th chunk is used as the test set. The first $5$ of the remaining sets are used as historical data to determine the deployed model's True Positive Rate (TPR) before deployment and the last final $5$ sets are used to iteratively update the deployed model in an Online fashion. These Online sets are susceptible to the Adam Outcome Performativity described in Sect. \ref{sect:ADAM_PERF}. At the end of each update, the model's new TPR is recorded. Once training is complete, a Mann-Whitney U test is used to determine if the difference in TPR achieved on the historical datasets are statistically different from the TPR achieved on the iterative (deployed) sets. For these experiments, reported results are the average over $25$ replicates and a $\delta=0.05$ was used to determine statistical significance.

\section{Training Machine Learning Models in Outcome Performative Settings}

Although it is not the focus of this paper, we do want demonstrate why training models in an Outcome Performative setting is different than in the traditional Supervised setting. For this, we will use the \verb|breast cancer| \cite{DATASET_BREAST_CANCER} dataset and the Simple Model of Outcome Performativity described in the paper. All results shown are average of 25 runs.

\begin{figure}
     \centering
     \begin{subfigure}[b]{0.3\textwidth}
         \centering
         \includegraphics[width=\textwidth]{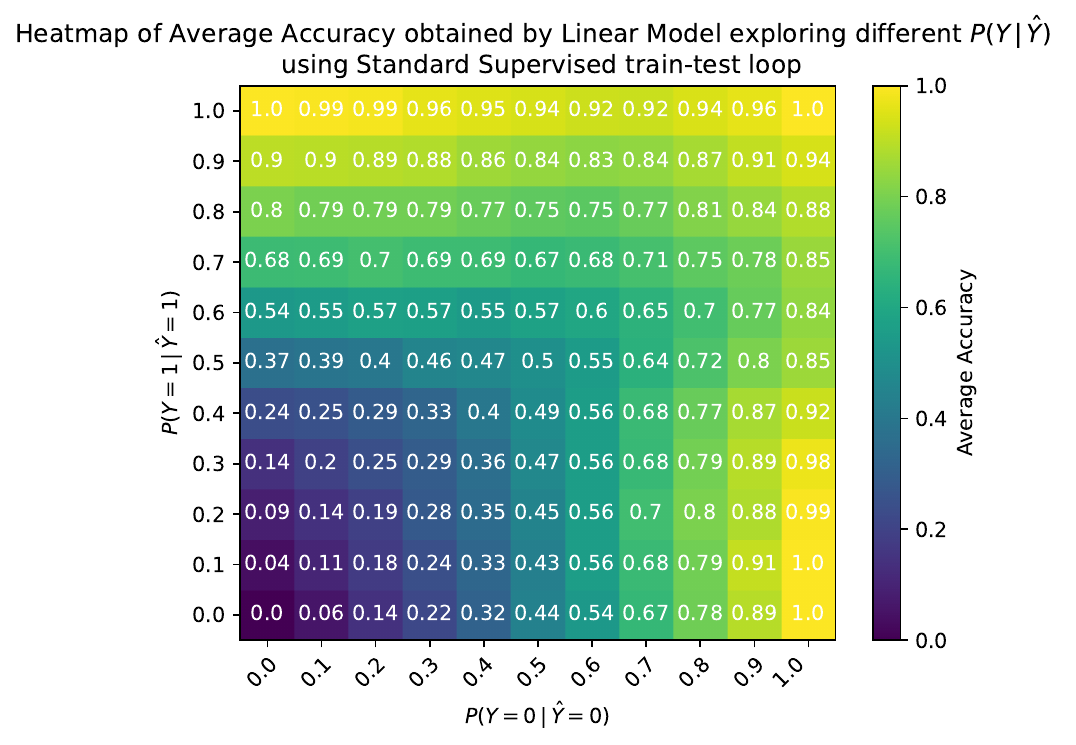}
         \caption{}
         \label{fig:OUTCOME_PERF_SUPERVISED}
     \end{subfigure}
     \hfill
     \begin{subfigure}[b]{0.3\textwidth}
         \centering
         \includegraphics[width=\textwidth]{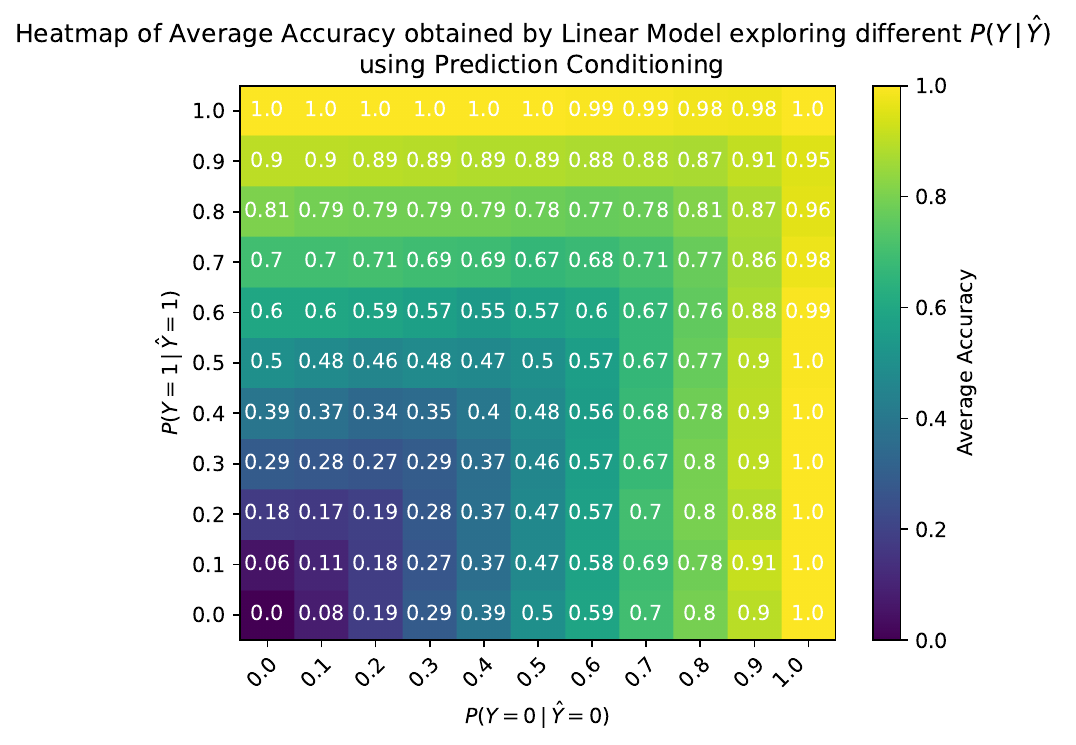}
         \caption{}
         \label{fig:OUTCOME_PERF_CONDITIONED}
     \end{subfigure}
     \hfill
     \begin{subfigure}[b]{0.25\textwidth}
         \centering
         \includegraphics[width=\textwidth]{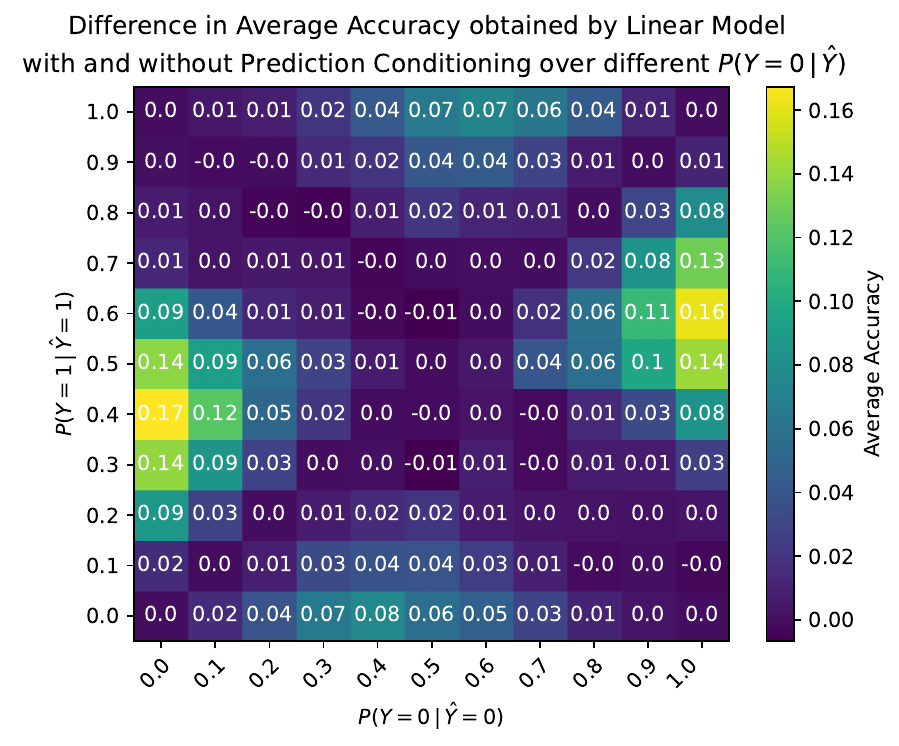}
         \caption{}
         \label{fig:OUTCOME_PERF_DIFF}
     \end{subfigure}
        \caption{Figure (a) shows the average accuracy achieved by a Linear Model (trained using standard Supervised Learning) over various parameters in the Simple Model of Outcome Performativity. Figure (b) plots the same using prediction conditioning whereby the Linear Model can predict the outcomes ($Y$) that will be observed given the prediction it makes $(\hat{Y}$). Figure (c) shows the difference in accuracy achieved by each model ($(b) - (a)$) with prediction conditioning performing better, particularly around areas of label imbalance in the original dataset.}
        \label{fig:LEARNING_WITH_OUTCOME_PERF}
\end{figure}

The first challenge in the Outcome Performative setting is gathering data. This has to come from historical data where a prediction $\hat{Y}$ has already been made and an outcome $Y$ observed. To simulate this in these demonstrations, we train a base classifier on the \verb|breast canncer| dataset, and use it's predictions to simulate outcomes for our real model $\theta$ to learn from. Assuming that we just use a standard Supervised training-test process, we would get results shown in Figure \ref{fig:OUTCOME_PERF_SUPERVISED}. Overall, the results are about what we expected, the accuracy of the model increases when the probability of encountering a label $P(Y \, | \, \hat{Y}) \to 1.0$. When $P(Y \, | \, \hat{Y}) \to 0.0$ the accuracy decreases as the model will never make a correct prediction. These results are unsurprising, but the model $\theta$ is far from achieving the maximum achievable accuracy of $\verb|max|(P(Y \, | \, \hat{Y} = 0), P(Y \, | \, \hat{Y} = 1))$. We can somewhat solve this problem by introducing \textit{prediction conditioning}. That is when the model can be conditioned on the prediction it would make in order to predict the final outcome it will observe. In practice, this is just adding an additional feature to the model $\hat{Y} = \{0, 1 \}$. 

This introduces a new problem where we must now choose between the probability scores produced for each class (e.g. $\theta_{\hat{y}=0}(x)$ vs. $\theta_{\hat{y}=1}(x)$ for a binary classification task). If we want to maximize accuracy as we did in the standard Supervised setting, we add a decision rule to determine our final prediction $\hat{y}$:

\begin{equation}
    \hat{y} = \begin{cases}
        0 & \verb|if | P_{\theta}(Y = 0 \, | \, \theta_{\hat{y}=0}(x) = 0) >  P_{\theta}(Y = 1 \, | \, \theta_{\hat{y}=1}(x) = 1)\\
        1 & \verb|else|
    \end{cases}
\end{equation}

where $P_{\theta}$ is the estimated probability score of outcome $Y$ when $\theta$ is conditioned on prediction $\hat{y}$. In short, this decision rule will predict $\hat{y} = 0$ if the estimated probability of observing outcome $y=0$ (given $\hat{y} = 0$) is greater than observing outcome $y=1$ (given $\hat{y} = 1$). The results of using a model with prediction conditioning is shown in Figure \ref{fig:OUTCOME_PERF_CONDITIONED}. Overall, the results look similar to that of the model without prediction conditioning, but if we observe Figure \ref{fig:OUTCOME_PERF_DIFF} which plots the difference in accuracy of the model with prediction conditioning $-$ the accuracy of the model without, we can see that that prediction conditioning can improve the accuracy of the model dramatically. For the \verb|breast cancer| dataset, the greatest benefits are at parameter ranges around that correlate with the data imbalances in the original dataset.

\begin{figure}
    \centering
    \includegraphics[width=0.5\linewidth]{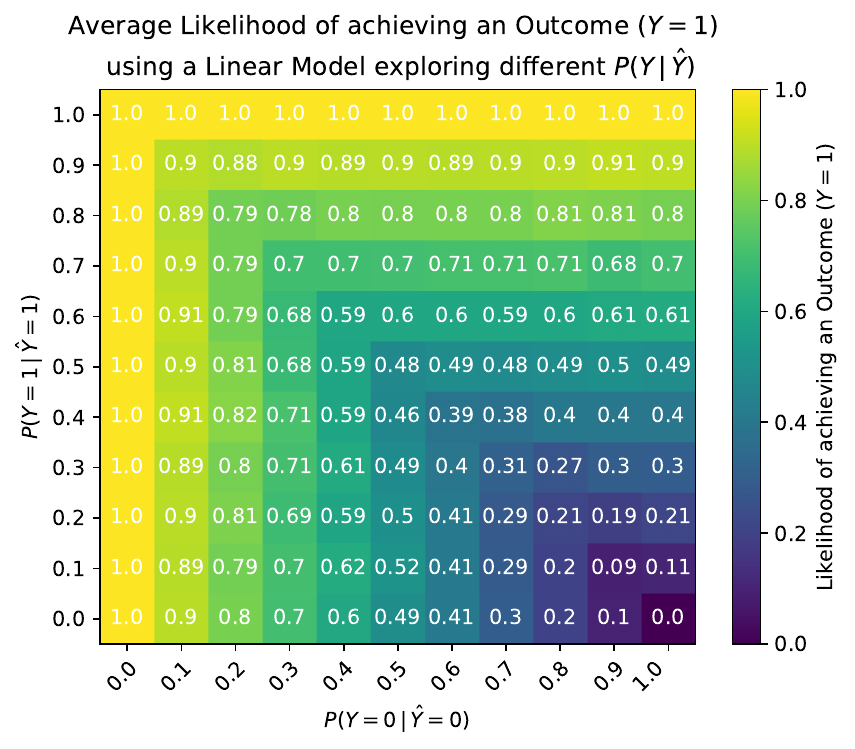}
    \caption{Figure showing how the same model used in Figure \ref{fig:OUTCOME_PERF_CONDITIONED} can be used to produce varying results over the same parameters. In this case, a simple change in the decision rule (from maximizing accuracy to maximizing observing an outcome of $y = 1$) produced the following figure. This is meant to highlight the difficulty and potential danger of naively deploying predictive models in settings that are Outcome Performative.}
    \label{fig:OUTCOME_MAXIMIZATION}
\end{figure}

Naively, these results imply that prediction conditioning is good, and sufficient for solving Outcome Performative problems. This may not always be the case. Yes, prediction conditioning does allow one to gain insight into how their predictions may affect observed outcomes, but it relies heavily on the decision rule you use. Is accuracy all you really care about, then the aforementioned decision rule works well. Consider the palliative care problem, we don't necessarily just want want to be accurate, we actually want to maximize an outcome (a patient's quality or life). If our deployed model is maximizing accuracy, it may over or under-prescribe interventional care. Fortunately, changing the model's behaviour is (in this abstract context) quite simple, we just need to change the decision-rule. We demonstrate that in Figure \ref{fig:OUTCOME_MAXIMIZATION} where we show how the same prediction conditioned model can be used to maximize different outcomes ($Y=1$ in this case) just by changing the decision rule to:

\begin{equation}
    \hat{y} = \begin{cases}
        0 & \verb|if | P_{\theta}(Y = 1 \, | \, \theta_{\hat{y}=0}(x) = 1) >  P_{\theta}(Y = 1 \, | \, \theta_{\hat{y}=1}(x) = 1)\\
        1 & \verb|else|
    \end{cases}
\end{equation}

The results clearly show that in the Outcome Performative setting, how you use your deployed model can have massive effect on the types of outputs you get. Before our model struggled to accurately predict outcomes in regions it is now achieving perfect scores in. This further acts as motivation for using OP-AB. Being aware of Outcome Performativity is the first step in understanding how to deploy a model in an Outcome Performative setting. These results also show how using prediction conditioning and different decision rules can produce vary different results, which further leads into ethical concerns that could arise given that these models will be deliberately steering outcomes towards results specified by their desired outcomes (i.e. decision rule).

\section{Applying OP-AB to Monitoring Tasks}

We introduced OP-AB as a "one-and-done" algorithm. That is, you give OP-AB $T$ instances to perform intervention testing on and the output of the algorithm is a "yes/no" answer to whether the setting is Outcome Performative. Practically, you may want to repeatedly evaluate a setting (perhaps the Outcome Performativeness of setting itself is dynamic or subject to Concept Drift). This would then be a monitoring task we you would have to repeatedly have to re-evaluate the Outcome Performativeness of your problem domain. Algorithm \ref{algo:OPAB_MONITORING} details how you might do this. Simply put, every $Z$ instances, you would dedicate $T$ instances to OP-AB and use the result to inform future decision making.

\begin{algorithm}[htbp]
\caption{Simple Pseudocode for using OP-AB in a monitoring task. The code is intentionally vague about what should be done when a positive test is found, and what is done with data batch $X$. This will be task specific.}
\label{algo:OPAB_MONITORING}
\begin{algorithmic}[1]
    \Require Data stream $X$, Test Frequency $Z$, Test Size $T$
    \State $x = X.next\_batch()$
    \State $T = 0$
    \While{$x \neq \varnothing$}
        \State $condition = T \verb| mod | Z = 0 \verb| and | OPAB(x, T)$
        \If{$condition$}
            \State Report Results
        \EndIf
        \State Perform operations on $x$
        \State $x = X.next()$
        \State $T += 1$
    \EndWhile
\end{algorithmic}
\end{algorithm}

\bibliographystyle{unsrt}  
\bibliography{references}  

\end{document}